\pdfoutput=1
\PassOptionsToPackage{dvipsnames, svgnames, x11names}{xcolor} 
\documentclass[11pt]{article}

\usepackage[preprint]{acl}

\usepackage{times}
\usepackage{latexsym}
\usepackage{booktabs}
\usepackage[T1]{fontenc}

\usepackage[utf8]{inputenc}

\usepackage{microtype}

\usepackage{inconsolata}

\usepackage{amsmath} 
\usepackage{array} 
\usepackage{multirow}
\usepackage{graphicx}
\usepackage{microtype}
\usepackage{inconsolata}

\usepackage{times}
\usepackage{latexsym}
\usepackage{adjustbox} 
\usepackage{graphicx}
\usepackage{listings}
\usepackage{todonotes}
\usepackage{subcaption}

\usepackage{url}

\usepackage{xcolor}
\usepackage{enumitem}
\PassOptionsToPackage{dvipsnames, svgnames, x11names}{xcolor}
\usepackage{tcolorbox} 

\usepackage{tabularx}
 
\usepackage{CJKutf8}

\usepackage{fvextra}
\DefineVerbatimEnvironment{myverb}{Verbatim}{breaklines,breakanywhere}

\author{
  \textbf{Bolei Ma\textsuperscript{$\ast$$\clubsuit$,$\heartsuit$}}~~~
    \textbf{Yina Yao\textsuperscript{$\ast$$\clubsuit$}}~~~
\textbf{Anna-Carolina Haensch\textsuperscript{$\clubsuit$,$\heartsuit$,$\diamondsuit$}}~~~
\vspace{7pt}
\\
\textsuperscript{$\clubsuit$}LMU Munich~~~
\textsuperscript{$\heartsuit$}Munich Center for Machine Learning\\\vspace{7pt}
\textsuperscript{$\diamondsuit$}University of Maryland, College Park
\\
  \small{$^\ast$Equal contributions.}\\
  \small{
    \textbf{Correspondence:} 
    \href{mailto:bolei.ma@lmu.de}{bolei.ma@lmu.de}
  }
}

\title{Capabilities and Evaluation Biases of Large Language Models in Classical Chinese Poetry Generation: A Case Study on Tang Poetry}

\begin{document}
\maketitle

\begin{abstract}

Large Language Models (LLMs) are increasingly applied to creative domains, yet their performance in classical Chinese poetry generation and evaluation remains poorly understood. We propose a three-step evaluation framework that combines computational metrics, LLM-as-a-judge assessment, and human expert validation. Using this framework, we evaluate six state-of-the-art LLMs across multiple dimensions of poetic quality, including themes, emotions, imagery, form, and style, in the context of Tang poetry (\begin{CJK}{UTF8}{gbsn}唐诗\end{CJK}) generation. Our analysis reveals a critical ``echo chamber'' effect: LLMs systematically overrate machine-generated poems that mimic statistical patterns yet fail strict prosodic rules, diverging significantly from human expert judgments. 
These findings underscore the limitations of using LLMs as standalone evaluators for culturally complex tasks, highlighting the necessity of hybrid human-model validation frameworks.\footnote{We release our code and generated poems at \url{https://github.com/boleima/Tang-Poetry}.}

\end{abstract}

\section{Introduction}

Large Language Models (LLMs) have recently shown promising performance in text generation, including applications to creative writing. Among these, classical Chinese poetry poses a unique challenge: it requires adherence to strict prosodic and tonal constraints while  achieving semantic richness, emotional resonance, and cultural authenticity \citep{Cai2008HowToReadChinesePoetry, He2012, Yang2018}. Tang poetry (\begin{CJK}{UTF8}{gbsn}唐诗\end{CJK}), in particular, is widely recognized as a pinnacle of Chinese literary tradition, and thus serves as an ideal testbed for evaluating AI creativity in a highly structured cultural domain.  

Despite the impressive fluency of modern LLMs, recent studies show that they still struggle with key aspects of poetry generation. Models often fail to maintain coherence across lines, exhibit limited originality in imagery, or collapse into reproducing memorized verses \citep{chakrabarty2023creative, gomez2023confederacy, chen2024evaluating}. These shortcomings suggest that poetry generation provides not only a challenging benchmark for natural language generation, but also a lens for examining the boundaries of machine creativity.  

A further challenge lies in evaluation. Traditional automatic metrics such as BLEU and ROUGE \citep{papineni2002bleu} capture surface-level similarity but fail to account for rhythm, imagery, or aesthetic quality \citep{Yang2018}. LLM-based evaluators (``LLM-as-a-judge'') have emerged as a promising alternative, offering richer semantic assessment \citep{liu-etal-2023-g, zhu2024eval}. However, concerns remain about systematic biases: models may inflate their own outputs, align with peers in an ``echo chamber'' fashion, or overlook critical dimensions such as prosodic accuracy \citep{clark2021all}. This raises fundamental questions about the reliability of automated evaluation in creative and culturally sensitive domains.

\begin{figure}
    \centering
    \includegraphics[width=1\linewidth]{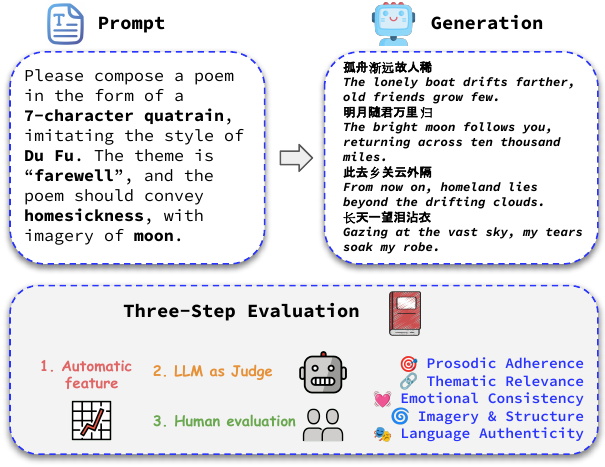}
    \caption{The basic framework of poetry generation and evaluation.}
    \label{fig:framework}
\end{figure}

In this work, we present the first systematic study of LLMs in generating and evaluating classical Chinese poetry, focusing on Tang poetry as a representative form. The basic research framework is shown in Figure \ref{fig:framework}. We let the models generate Tang poems giving instructions of different dimensions including form, poet, theme, emotion and imagery. We then introduce a three-step evaluation framework that integrates (1) computational feature extraction, (2) LLM-as-a-judge evaluation, and (3) human expert validation. Applying this framework to six state-of-the-art LLMs, we uncover distinct performance tiers in generation quality and reveal systematic evaluation biases that deviate from human judgments.  

Our contributions are threefold:  

\vspace{5pt}
\begin{itemize}[leftmargin=*,nolistsep]
    \item We propose a multi-dimensional three-step framework for evaluating LLM-generated Chinese poetry.
    \item We conduct the first empirical study of self- and cross-evaluation biases among LLMs in the classical Chinese poetry domain, identifying ``echo chamber'' effects in automated assessment.  
    \item We provide comparative benchmarks across six LLMs, highlighting strengths, weaknesses, and the necessity of hybrid collaboration of human and AI validation in evaluating creative AI.
\end{itemize}

\section{Related Work}
\label{sec:related}
\paragraph{Chinese Classical Poetry as an NLP Generation Task.}  
Tang poetry is characterized by strict metrical rules and tonal patterns, making it a demanding test case for text generation. Beyond formal correctness, effective poetry must convey imagery, emotion, and cultural resonance \citep{Cai2008HowToReadChinesePoetry, Yang2018,ma-etal-2023-yu,huang-shen-2025-poembert,zou-2025-bipro}. These qualities distinguish poetry generation from standard text generation tasks and require evaluation frameworks that capture both structural and aesthetic dimensions.  

\paragraph{LLMs for Poetry Generation.} 
These are existing works on LLMs for generating poetry in Languages such as French \cite{HamalainenAP22}, Greek \cite{chatzikyriakidis-natsina-2025-poetry}, and Russion \cite{koziev-fenogenova-2025-generation}. 
While early work on Chinese poetry generation explored template- or retrieval-based systems \citep{He2012}, recent advances in LLMs have greatly improved fluency and stylistic control \citep{Yu2024, liu2025large}. There are also LLMs trained for the poetry tasks \cite{ijcai2025p927}. Nonetheless, studies have documented persistent challenges: models often produce repetitive or collage-like outputs, lack originality in imagery, and fail to consistently obey prosodic rules \citep{chakrabarty2023creative, chen2024evaluating}. These limitations motivate systematic evaluation of current models in highly constrained creative tasks such as Tang poetry.  

\paragraph{Evaluation of Generated Poetry.}  
Automatic evaluation remains a bottleneck. Metrics like BLEU and ROUGE \citep{papineni2002bleu} are ill-suited to capturing the cultural and aesthetic value of poetry \citep{Yang2018}. LLM-as-a-judge approaches provide richer assessments \citep{liu-etal-2023-g, zhu2024eval}, and multi-dimensional evaluation frameworks have been proposed for creative generation \citep{guo2019jiuge, deng2024turing}. Yet, recent work highlights systematic biases in LLM evaluations, including inflated self-assessments and convergence toward flawed standards \citep{clark2021all}. This motivates integrating automated metrics with human expert validation to ensure reliability in evaluating poetry.

\section{Experimental Design and Methods}

This section integrates the experimental design and analytical methods for generating and evaluating the Tang poems.  
As shown in Figure \ref{fig:framework}, the overall methodology involves large-scale generation, and a three-step evaluation framework (computational features, LLM cross-evaluation, and human expert validation).

\subsection{Generation}
\label{sec:generation}
\paragraph{Models.}  
We select six open-source instruction-tuned LLMs as generators and evaluators: DeepSeek-V2-Lite-Chat \citep{deepseek}, Qwen2.5-7B-Instruct \citep{qwen}, GLM-4-9B-Chat-HF \citep{glm}, Mistral-7B-Instruct-v0.3 \citep{mistral}, Baichuan2-7B-Chat \citep{baichuan}, Gemma-2-9B-it \citep{gemma}.

\paragraph{Poetry Dimensions.} We cover five important dimensions for Tang poetry: poetic form, target poet style, theme, emotion, and imagery (Table~\ref{tab:poetry_dimensions}). For each dimension, we list representative example elements. 
For each model, we generate a controlled cross sample set (15,000 poems in total, around 2,500 per model), covering all five dimensions and their elements.

\begin{table}[htbp]
\centering
\small
\setlength{\tabcolsep}{1pt}
\begin{tabular}{|l|l|l|}
\hline
\textbf{Dim.} & \textbf{Specific Elements} \\
\hline
Form & 5/7-character quatrains/regulated verse \\
\hline
Poet & Li Bai, Du Fu, Bai Juyi, Wang Wei, Li Shangyin \\
\hline
Theme & landscape, homesickness, nostalgic, pastoral, parting \\
\hline
Emotion & sadness, tranquility, boldness, romance, joy \\
\hline
Imagery & wind, flowers, willows, moon, geese \\
\hline
\end{tabular}
\caption{Poetry dimensions and elements for generation. The original Chinese version is in Appendix \ref{sec:dim_zh}.}
\label{tab:poetry_dimensions}
\end{table}

\paragraph{Prompt.} Generation is guided by explicit prompts specifying form and imitation target, with temperature set to $T=0.4$ for diversity. An example prompt is shown in Appendix \ref{sec:prompt}.

\subsection{Evaluation}

We propose a three-step evaluation framework (Figure~\ref{fig:framework}) designed to assess both the generative quality and evaluation biases of LLMs in Tang poetry. The framework integrates (i) automated feature analysis, (ii) LLM-as-judge cross-assessment, and (iii) human expert validation.

\subsubsection{Evaluation Pipeline}

Evaluation proceeds in three parts. First, automated feature extraction provides objective measures of lexical and semantic properties of generated poems. 

Second, in the LLM-as-judge stage, a $6 \times 6$ evaluation matrix is constructed, where each model acts both as generator and evaluator, scoring all generated samples. This setup enables simultaneous observation of self-assessment and peer-assessment behaviors. Each evaluation is conducted along five dimensions: 
\vspace{10pt}
\begin{itemize}[leftmargin=*, nolistsep]
\item  \textbf{\textit{Prosodic Adherence:}} Assesses the harmony of sounds, correctness of tonal patterns, use of rhyme, and overall rhythmic quality. 

\item \textbf{\textit{Thematic Relevance:}} Judges the depth, clarity, and accuracy with which the poem expressed its designated theme. 

\item \textbf{\textit{Emotional Consistency:}} Evaluates the authenticity and resonance of the emotion conveyed, ensuring it aligned with the specified emotional tone. 

\item \textbf{\textit{Imagery and Structure:}} Assesses the vividness and originality of the poem's imagery, as well as its logical coherence and structural integrity. 

\item \textbf{\textit{Language Authenticity:}} Judges the precision, conciseness, and authenticity of the language used.
\end{itemize}
\vspace{10pt}
These five dimensions were adopted by synthesizing the core structural and aesthetic criteria defined in classical Chinese literary theory \cite{Liu1962ArtOfChinesePoetry, Cai2008HowToReadChinesePoetry}. Scores range from 1 to 5.

Finally, in the human validation stage (Step~3), the top three models (ranked by peer evaluation, excluding self-scores) are selected. From each, 50 poems (150 total) are sampled and anonymized. Experts in classical Chinese literature evaluate these poems using the same five dimensions, providing an external calibration of automated evaluation.

\subsubsection{Evaluation Methods (three-step analysis)}

We detail the analytical methods used in each step.

\paragraph{Step 1: Computational feature extraction.}

We employ three categories of objective metrics to characterize lexical and semantic properties of generated poems.\footnote{These metrics capture diversity and stylistic tendencies but do not directly measure literary quality; they must be interpreted in conjunction with step~2/3.}  

\begin{itemize}[leftmargin=*]

\item[1)] \textbf{TF-IDF.} We compute TF-IDF to extract lexical fingerprints of poems and models, quantifying characteristic vocabulary usage. For poem $d \in D$ and term $t$:

{\small
\begin{equation}
\mathrm{TF\!-\!IDF}(t,d,D)=\mathrm{TF}(t,d)\cdot\log\frac{|D|}{|\{d' \in D:t\in d'\}|}
\label{eq:tfidf}
\end{equation}
}

\noindent enabling comparison of lexical tendencies across models \citep{sparck1972statistical,manning1999foundations}.

\item[2)] \textbf{Shannon entropy.} To assess lexical diversity, we compute Shannon entropy:

\begin{equation}
H(X)=-\sum_{i} p(x_i)\log_2 p(x_i)
\label{eq:entropy}
\end{equation}

\noindent Distributions are compared via one-way ANOVA, with Tukey HSD post-hoc tests when significant, distinguishing between ``formulaic'' and ``diverse'' generative styles.

\item[3)] \textbf{Semantic similarity.} To measure thematic relevance and internal coherence, we embed sentences/poems using Chinese sentence embeddings and compute cosine similarity:

\begin{equation}
\mathrm{sim}(A,B)=\frac{A\cdot B}{\|A\|\|B\|}
\end{equation}

\noindent This is applied to poem-prompt alignment and average line-to-line coherence \citep{reimers2019sentence}.

\end{itemize}

\paragraph{Step 2: Descriptive analysis of LLM-as-judge scores. }

For the $6\times 6$ matrix of cross-assessments, we conduct descriptive and inferential analyses:

\begin{itemize}[leftmargin=*]
  \item \textbf{Generator effect.} Column averages yield consensus rankings of generators.  
  \item \textbf{Evaluator effect.} Row averages capture evaluator temperament (leniency vs.\ harshness).  
\end{itemize}

This analysis provides both performance ranking and diagnostic insight into evaluation biases.

\paragraph{Step 3: Human-AI alignment.}

To measure agreement between automated and human evaluation, we compute Spearman rank correlation between expert scores and mean LLM scores:

\begin{equation}
\rho = 1 - \frac{6\sum d_i^2}{n(n^2-1)}
\end{equation}

\noindent where $d_i$ is the rank difference for poem $i$, and $n$ is the sample size. High $\rho$ indicates stronger alignment, while low values reveal systematic blind spots (e.g., insensitivity to metrical violations).

\textbf{Sample and statistical strategy.}  
The human evaluation set comprises 150 anonymized poems (50 per model from the top three peer-ranked models). Each poem is rated by a single expert.

\section{Experiments and Results}
This chapter presents the analysis results of the comparative study, structured according to the three-step evaluation framework. 

\subsection{Computational Feature Analysis}
The first analysis employs quantitative methods to objectively characterize the textual properties of the generated poems. Three metrics are considered: lexical fingerprinting (TF-IDF), lexical diversity (Shannon entropy), and semantic coherence (embedding-based similarity). 

\subsubsection{Lexical Fingerprinting}
To identify the characteristic vocabulary of each generative model, we applied the TF-IDF weighting defined in Eq.~\ref{eq:tfidf}. By calculating scores for all terms across the corpus, a distinct ``lexical fingerprint'' was derived for each of the six models.

\begin{figure}
    \centering
    \includegraphics[width=1.05\linewidth]{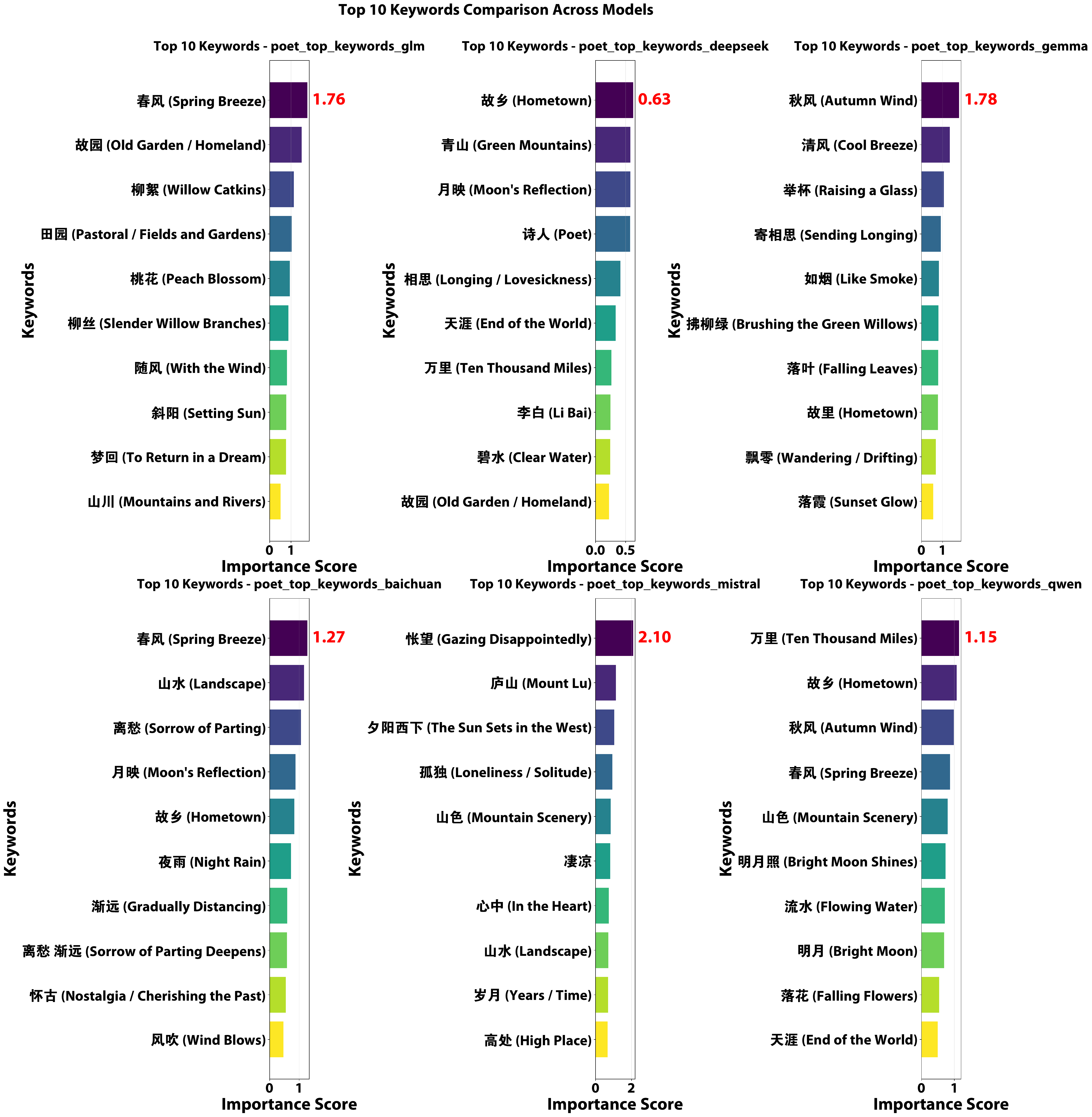}
    \caption{Top 10 keywords for all models, with highest TF-IDF scores marked red for each model.}
    \label{fig:top15}
\end{figure}

\textbf{Shared classical lexicon captured the foundational thematic elements of the genre.}  
A comparative analysis of the top 10 keywords with the highest TF-IDF scores for each model (see Figure \ref{fig:top15}) shows that all models successfully reproduce the core lexicon of classical Chinese poetry. Traditional poetic imagery such as \begin{CJK}{UTF8}{gbsn}春风\end{CJK} (``spring breeze''), \begin{CJK}{UTF8}{gbsn}明月\end{CJK} (``bright moon''), \begin{CJK}{UTF8}{gbsn}青山\end{CJK} (``green mountains''), and \begin{CJK}{UTF8}{gbsn}故乡\end{CJK} (``hometown'') appears frequently across models. This common vocabulary suggests that all generators captured the foundational thematic elements of the genre.

\textbf{Distinct lexical preferences reveal model-specific styles.}  
Closer inspection, however, reveals clear stylistic divergences. Qwen and DeepSeek, for example, show a marked preference for expansive landscape expressions, with terms such as \begin{CJK}{UTF8}{gbsn}万里\end{CJK} (``ten thousand miles'') and \begin{CJK}{UTF8}{gbsn}天下\end{CJK} (``all under heaven'') scoring prominently. In contrast, Baichuan, Gemma, and Mistral lean toward more personal and emotional expressions, frequently using words such as \begin{CJK}{UTF8}{gbsn}离愁\end{CJK} (``sorrow of parting''), \begin{CJK}{UTF8}{gbsn}萧瑟\end{CJK} (``desolate''), and \begin{CJK}{UTF8}{gbsn}孤独\end{CJK} (``loneliness''). By comparison, GLM favors more joyful and serene vocabulary, such as \begin{CJK}{UTF8}{gbsn}春风\end{CJK} (``spring breeze'') and \begin{CJK}{UTF8}{gbsn}悠然\end{CJK} (``leisurely''). These differences suggest that models have internalized stylistic patterns from their training corpora.

\subsubsection{Lexical Diversity and Predictability}
Lexical diversity and unpredictability were quantified using Shannon entropy ($H$), defined in Eq.~\ref{eq:entropy}. Entropy measures the average uncertainty of the word distribution in a text: higher values indicate richer lexical variety and lower predictability.

\begin{figure}[htbp]
    \centering
    \includegraphics[width=1\linewidth]{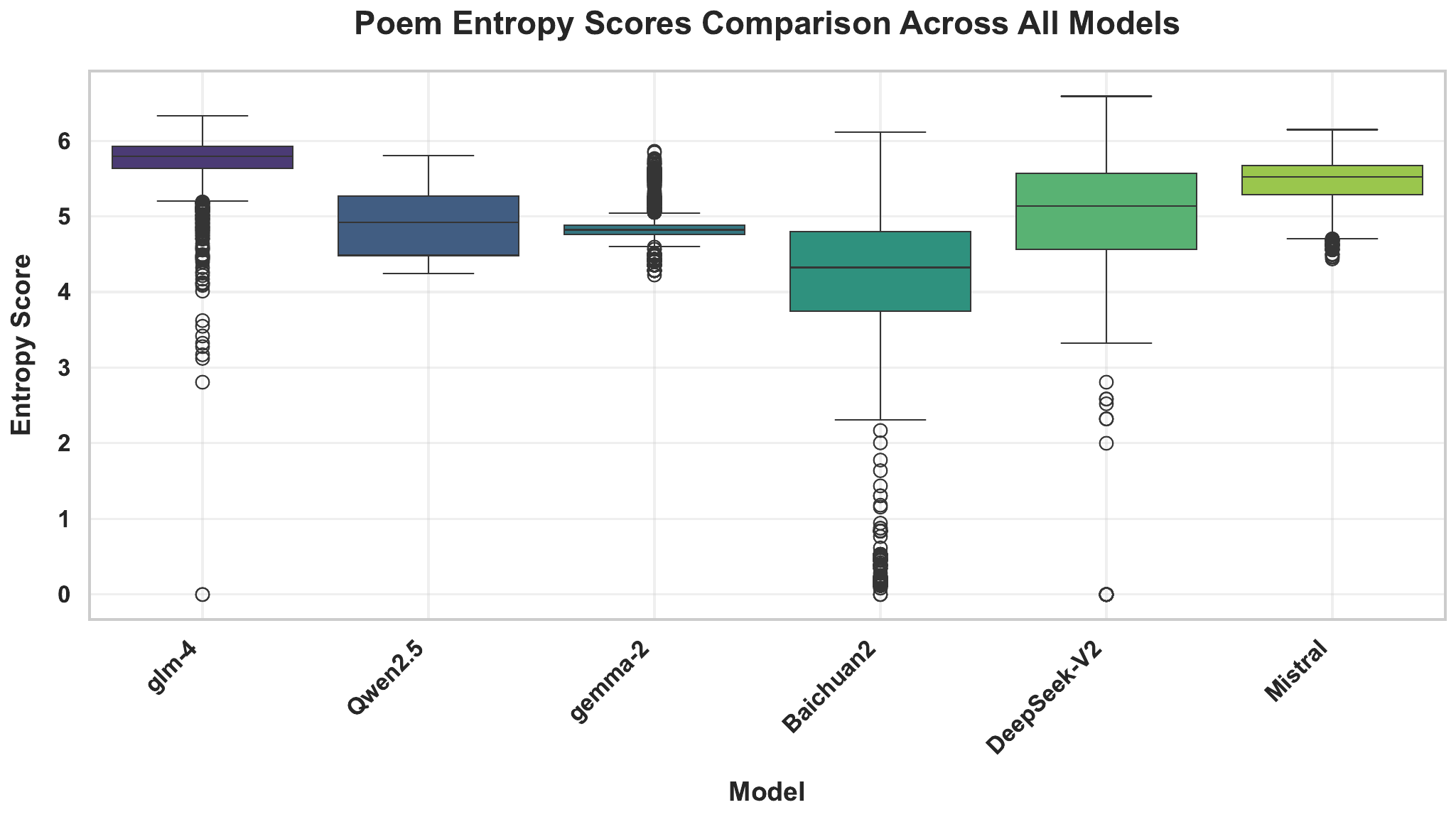}
    \caption{Entropy scores across models.} 
    \label{fig:entropy}
\end{figure}

The analysis reveals \textbf{three distinct performance tiers among the six models} (see Figure \ref{fig:entropy}). 
\textbf{Class 1 (High entropy): GLM and Mistral.} Both produce poems with the highest lexical diversity. 
\textbf{Class 2 (Medium entropy): DeepSeek, Qwen, and Gemma.} These models form a central cluster. 
\textbf{Class 3 (Low-entropy outlier): Baichuan}, which consistently yields the lowest lexical diversity.  
To formally validate these observations, a one-way ANOVA was conducted, yielding a highly significant result ($F(5, 14994) = 1986.6, p < .001$). Post-hoc comparisons with Tukey's HSD (see Table \ref{tab:tukey_hsd_entropy}) confirm the tiered structure. Results confirm three entropy-based classes: Baichuan as an outlier, a mid-tier cluster (DeepSeek, Qwen, Gemma), and a high-entropy group (Mistral, GLM).

\begin{table}[!htbp]
\centering
\tiny
\setlength\tabcolsep{1pt}
\renewcommand\arraystretch{0.85}
\begin{tabular}{llcccc}
\toprule
\textbf{Model Group 1} & \textbf{Model Group 2} & \textbf{Mean Diff.} & \textbf{p-adj} & \textbf{Reject $H_0$} & \textbf{Conclusion} \\
\midrule
Baichuan & Other Models & > 1.03 & <.001 & True & Stat. Unique (Class 3) \\
DeepSeek & Qwen             & 0.0197 & 0.936 & False & Stat. Indistinguishable \\
DeepSeek & Gemma            & -0.1865 & <.001 & True & Sig. Different \\
Mistral  & GLM              & 0.2652 & <.001 & True & Sig. Different \\
Mid-Tier Models & High-Tier Models & > 0.43 & <.001 & True & Sig. Different \\
\bottomrule
\end{tabular}
\caption{Pairwise comparison of mean entropy scores using Tukey's HSD.} 
\label{tab:tukey_hsd_entropy}
\end{table}

\begin{figure}[!htbp]
    \centering
    \includegraphics[width=1\linewidth]{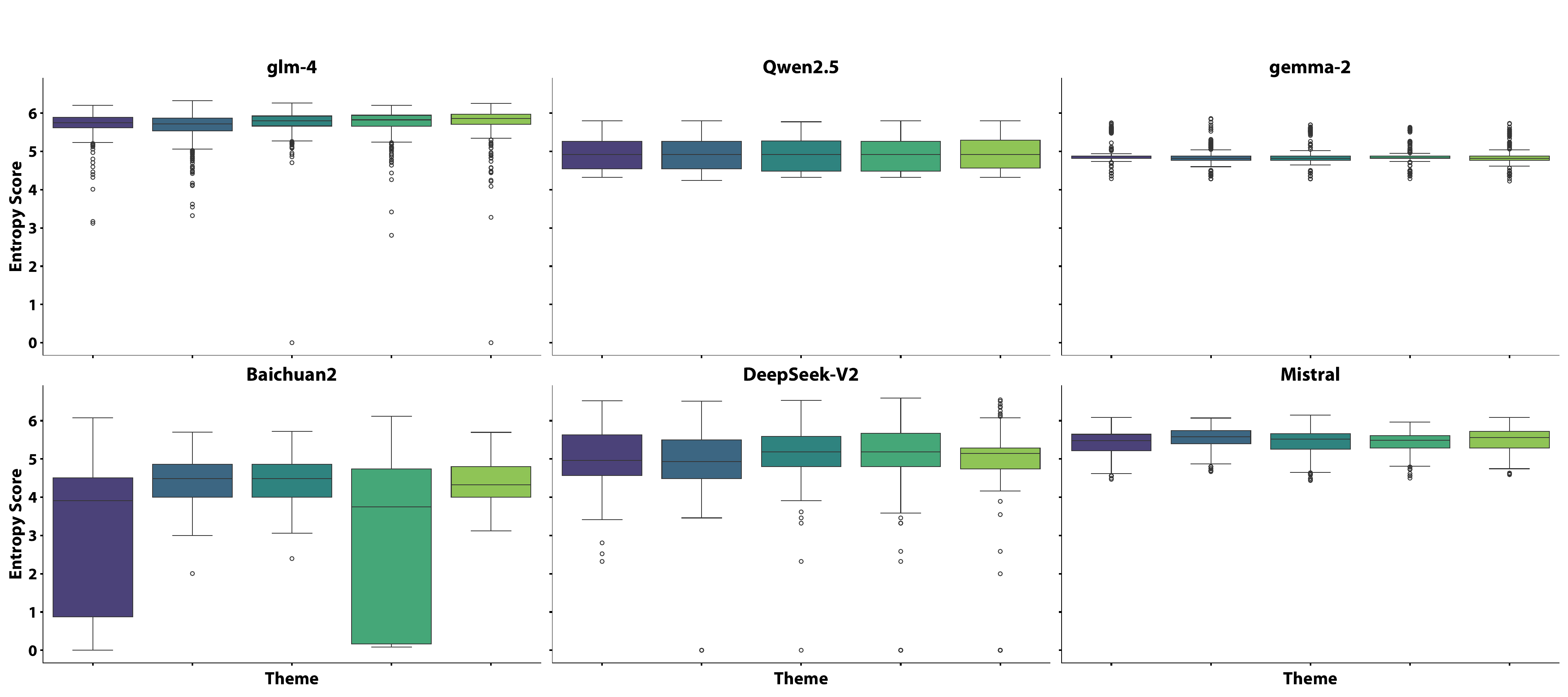}
    \caption{Entropy scores across models by theme.}
    \label{fig:entropy by theme}
\end{figure}

\textbf{The strict prosodic rules of regulated poetry naturally constrain lexical choices, reducing entropy.} Hence, Baichuan's low entropy may reflect either (a) genuine lexical limitation and repetitiveness, or (b) stronger mastery of poetic form. Conversely, the high entropy of GLM and Mistral could signal creative flexibility or, alternatively, a tendency to sacrifice formal correctness for lexical variety. 
Further analysis of entropy by theme (Figure \ref{fig:entropy by theme}) provides additional nuance. Baichuan's entropy decreases sharply when tasked with thematically specific prompts (e.g., \begin{CJK}{UTF8}{gbsn}怀古\end{CJK} ``Nostalgic''), indicating vocabulary restriction and repeated reliance on signature terms such as \begin{CJK}{UTF8}{gbsn}万里\end{CJK} (``ten thousand miles``). Combined with TF-IDF results, this suggests \textbf{a template-driven generation strategy, in which iconic words are repeatedly reused rather than flexibly adapted to context.}

\subsubsection{Semantic and Cultural Coherence}
\label{sec:semantic and cultural coherence}

To address the ambiguities left by the entropy analysis and to move beyond purely lexical metrics, we conducted a semantic analysis using the \href{https://huggingface.co/shibing624/text2vec-base-chinese}{\texttt{shibing624/text2vec-base-chinese}} sentence embedding model. 

\begin{figure}[!htbp]
    \centering
    \includegraphics[width=1\linewidth]{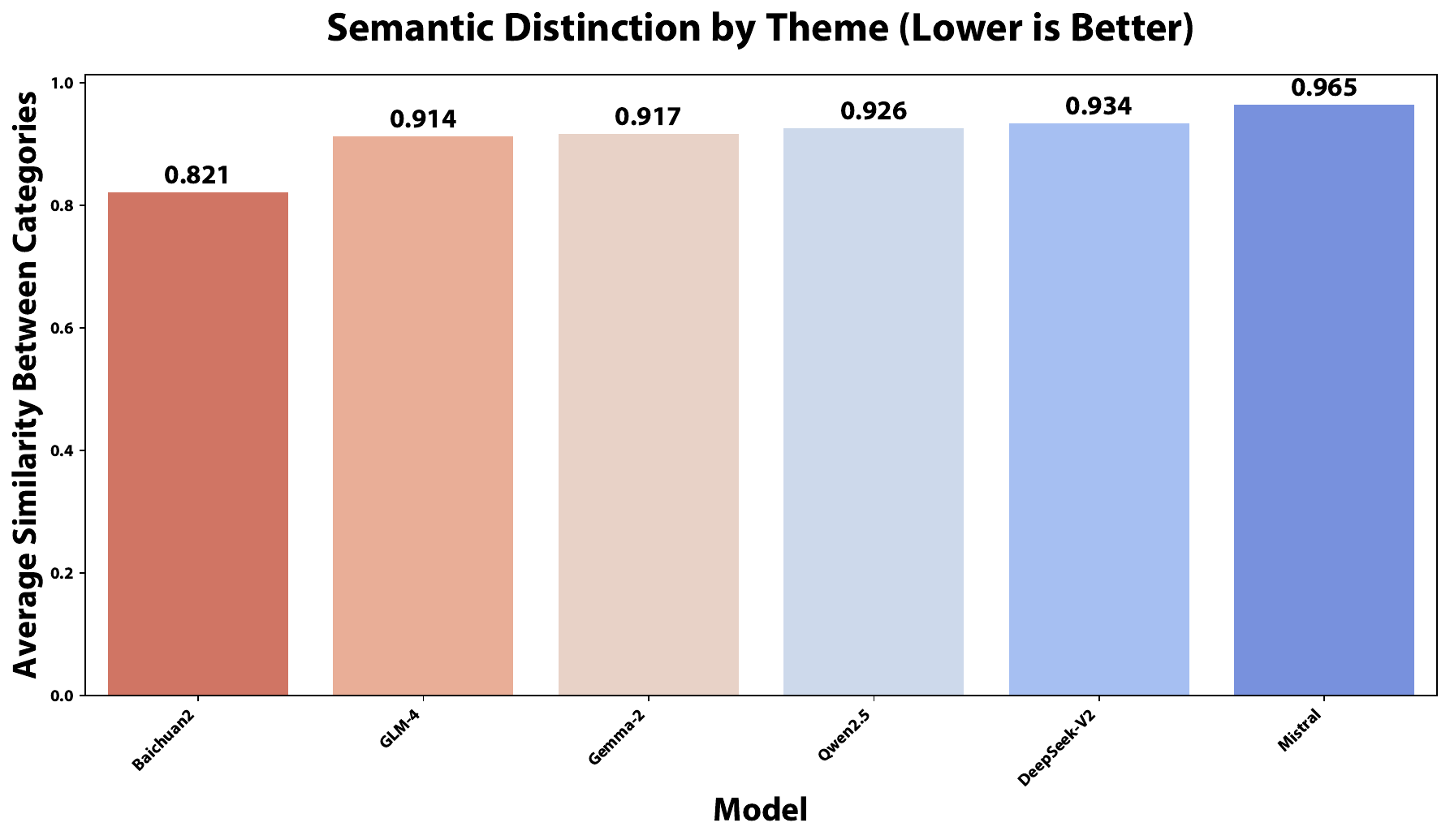}
    \caption{Semantic distinction by theme. }
    \label{fig:semantic_theme}
\end{figure}

\textbf{Semantic Thema Distinction: Lexical diversity does not guarantee semantic or stylistic richness.} 
This analysis evaluates how well models generate semantically distinct outputs across creative prompts. Lower similarity scores indicate stronger separation. The results (Figure \ref{fig:semantic_theme}) show 
that models do not separate themes well. 
Baichuan performs better with lowest similarity score (0.821). This aligns with its low entropy: concentrated thematic focus can naturally yield more predictable vocabulary, suggesting that its low entropy reflects thematic control rather than lexical weakness. 
Even high-entropy models like GLM and Mistral fail to convert lexical variety into semantic nuance. This reinforces the entropy Paradox: lexical diversity does not guarantee semantic or stylistic richness.

\begin{figure}[!htbp]
    \centering
    \includegraphics[width=1\linewidth]{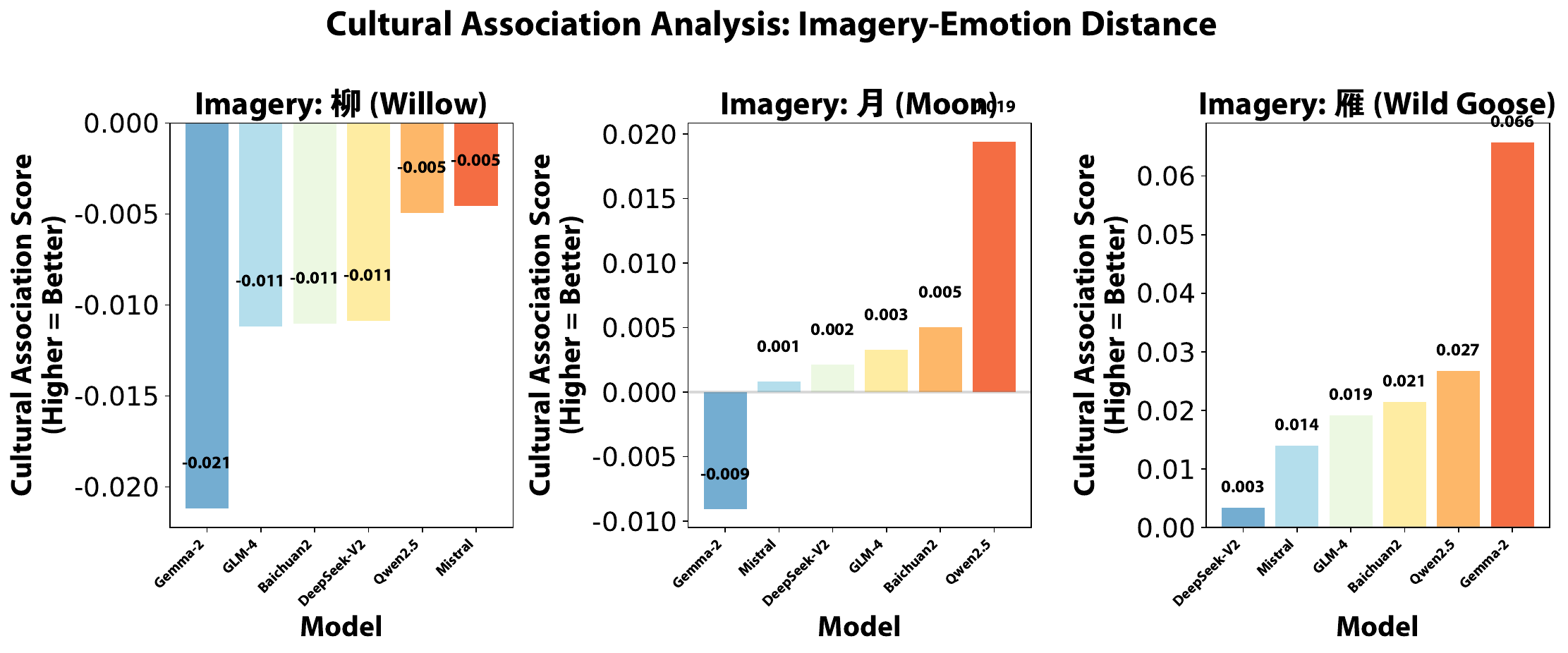}
    \caption{Imagery-Emotion Association in cultural association scores.} 
    \label{fig:IE}
\end{figure}

\textbf{Generated poems reveal cultural blind spots.}
This analysis examines the models' grasp of classical Chinese ``imagery-emotion'' pairings (Figure \ref{fig:IE}). They reliably reproduced canonical links such as Moon -> Tranquility and Wild Goose -> Sadness, with Gemma performing especially well on the latter. Yet every model failed to capture the quintessential Willow -> Sadness association, a cornerstone of Chinese poetic symbolism where willow branches signify parting and melancholy.\footnote{The willow (\begin{CJK}{UTF8}{gbsn}柳, liǔ\end{CJK}) is a canonical cultural allusion in Chinese poetry for parting and melancholy (due to the homophony with \begin{CJK}{UTF8}{gbsn}留, liú\end{CJK}, meaning to stay) \cite{Peng_Cao_Shi_Cao_2024}. For the target genre and theme, this is a cornerstone of poetic symbolism. In Chinese poetry, there is canonical links between the imagery and emotions, such as Moon -> Tranquility, Wild Goose -> Sadness, Willow -> Sadness, as detailed in Classical Chinese studies such as \citet{Cai2008HowToReadChinesePoetry}.} This omission highlights a critical deficit in cultural association fidelity that undermines the authenticity of the generated poems. The gap may stem from sparse representation of this motif in broad training data or from deeper architectural limits in modeling nuanced metaphorical relationships. 

\begin{figure}[!htbp]
    \centering
    \begin{subfigure}[t]{0.495\linewidth}
        \centering
        \includegraphics[width=\linewidth]{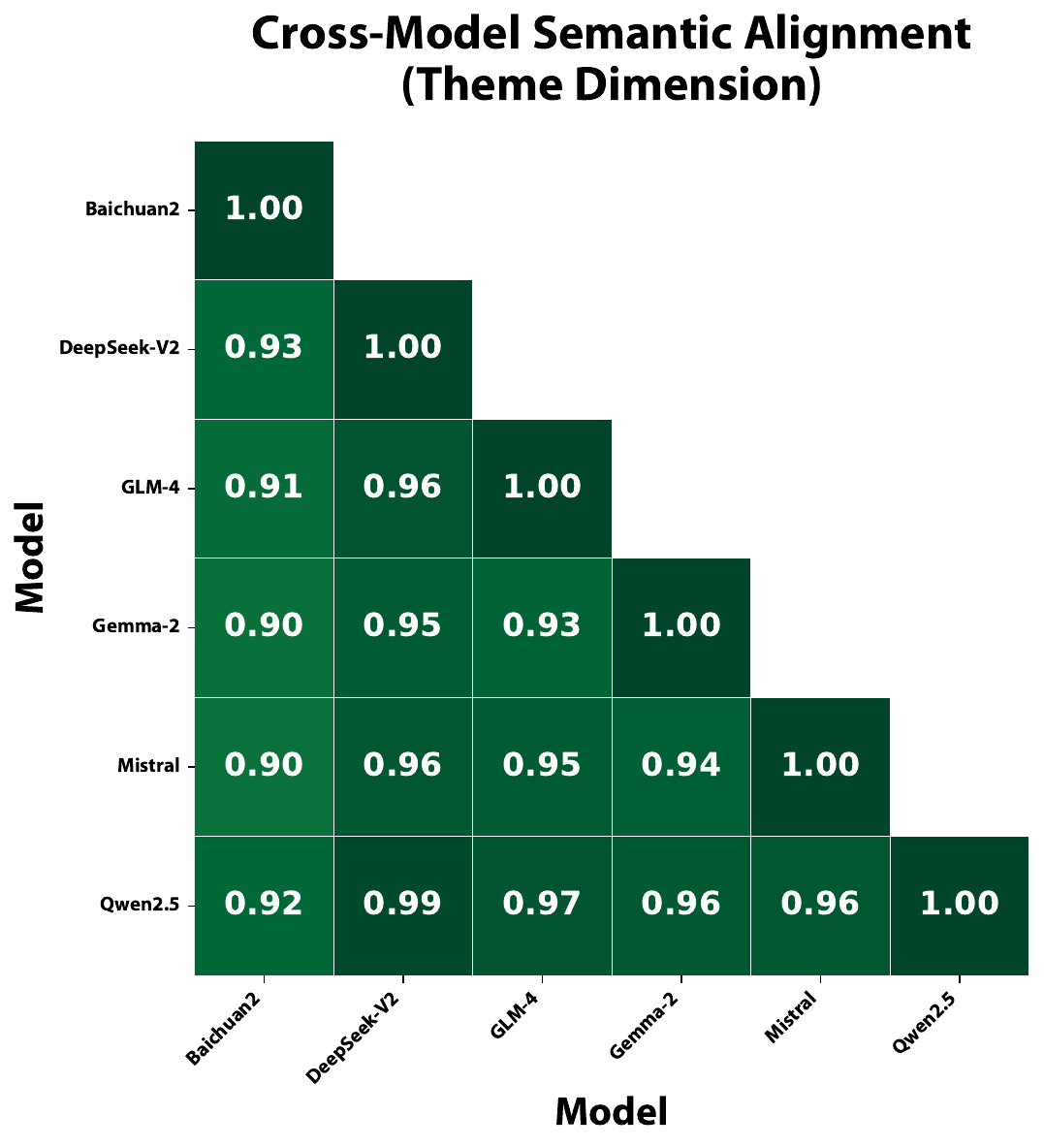}
        \caption{By theme}
        \label{fig:SAT}
    \end{subfigure}\hfill
    \begin{subfigure}[t]{0.495\linewidth}
        \centering
        \includegraphics[width=\linewidth]{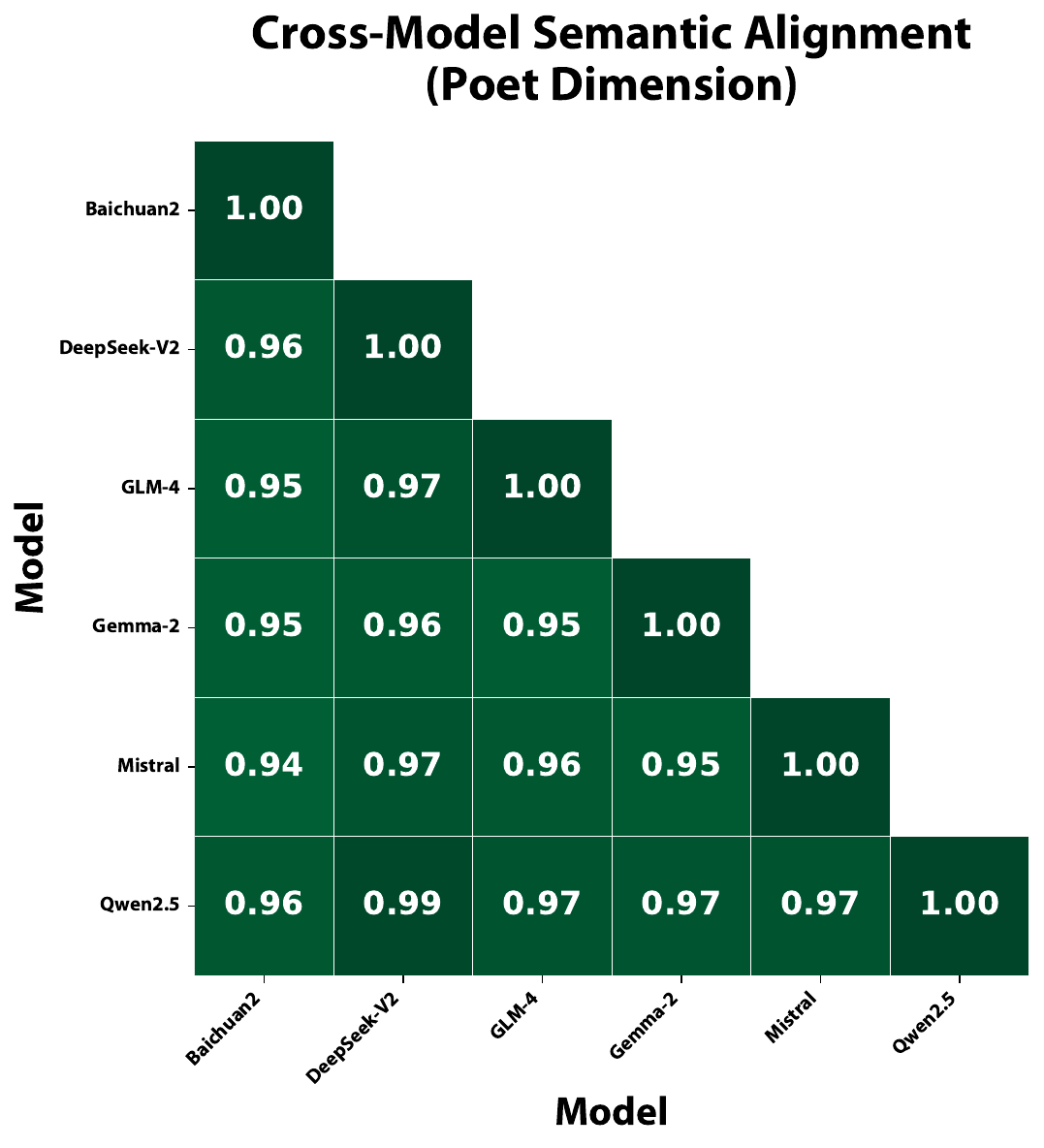}
        \caption{By poet}
        \label{fig:SAP}
    \end{subfigure}
    \caption{Cross-model semantic alignment.}
    \label{fig:SA}
\end{figure}

\begin{figure}[htbp]
    \centering
    \includegraphics[width=1\linewidth]{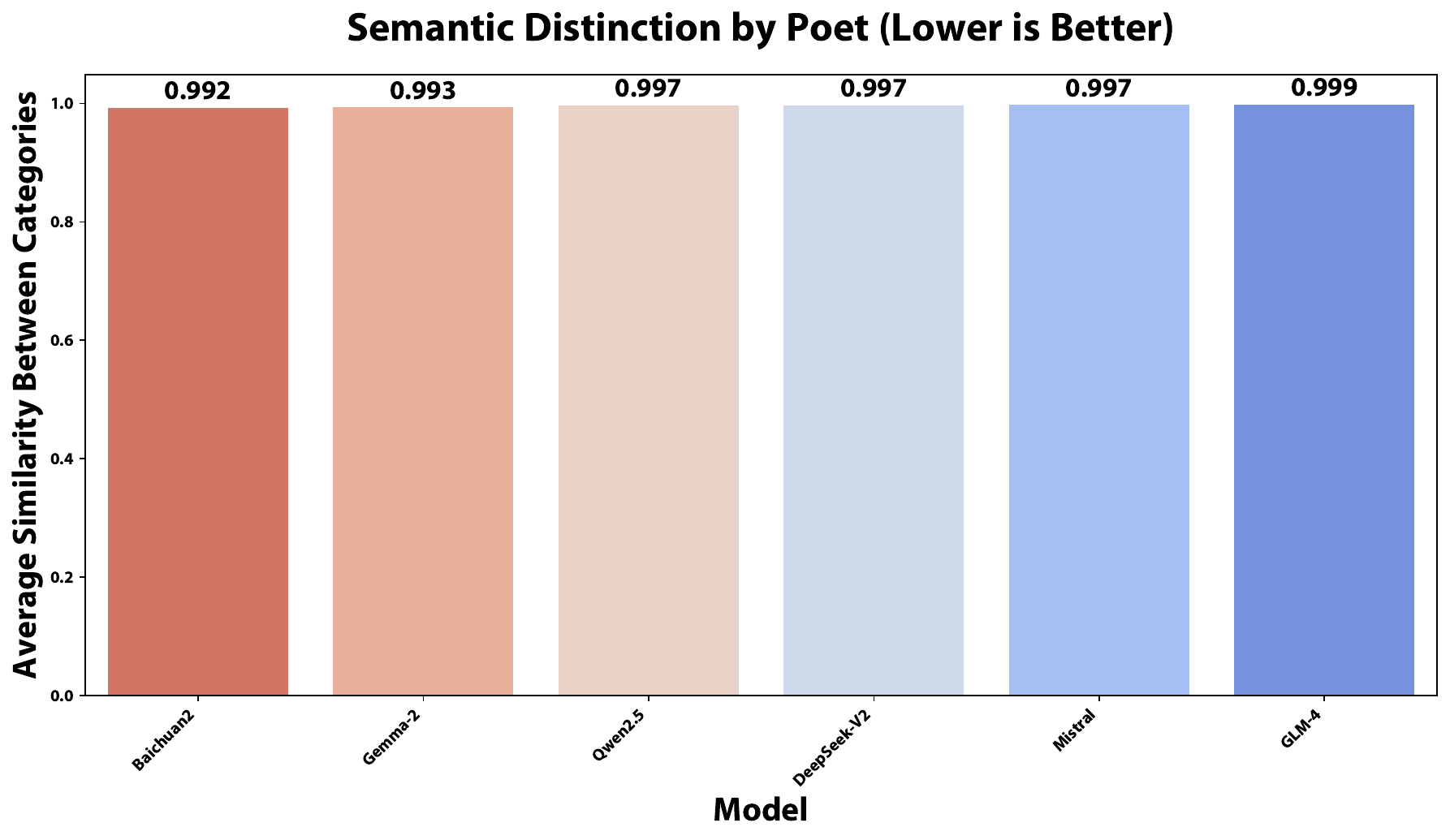}
    \caption{Semantic distinction by poets. }
    \label{fig:semantic_poets}
\end{figure}

\textbf{Models exhibit strong semantic convergence.}
We measured cross-model semantic similarity (Figures \ref{fig:SAT} and \ref{fig:SAP}) and found mean scores above 0.94, revealing a shared representation of poetic meaning likely rooted in overlapping training data. Despite noticeable lexical variation, the underlying semantics are strikingly uniform. 
We clarify that the alignment paradox refers to the contradictory relationship between a model's internal semantic knowledge and its actual generative output. 
We further show the average similarity by poets in Figure \ref{fig:semantic_poets} (the lower, the more distinct). Similar to the findings from Figure \ref{fig:semantic_theme}, models
do not separate poems generated for different poets well. 
This suggests that while prompting different models with different poets (e.g., \begin{CJK}{UTF8}{gbsn}李白\end{CJK} ``Li Bai'' vs. \begin{CJK}{UTF8}{gbsn}杜甫\end{CJK} ``Du Fu''), they struggle to manifest these differences in the practice when generating poems. They are essentially trapped in a shared semantic space where they can describe stylistic differences in theory but produce outputs that are semantically indistinguishable from one another.

\subsection{LLMs-as-judges Evaluation}

\begin{figure}[!htbp]
    \centering
    \includegraphics[width=1\linewidth]{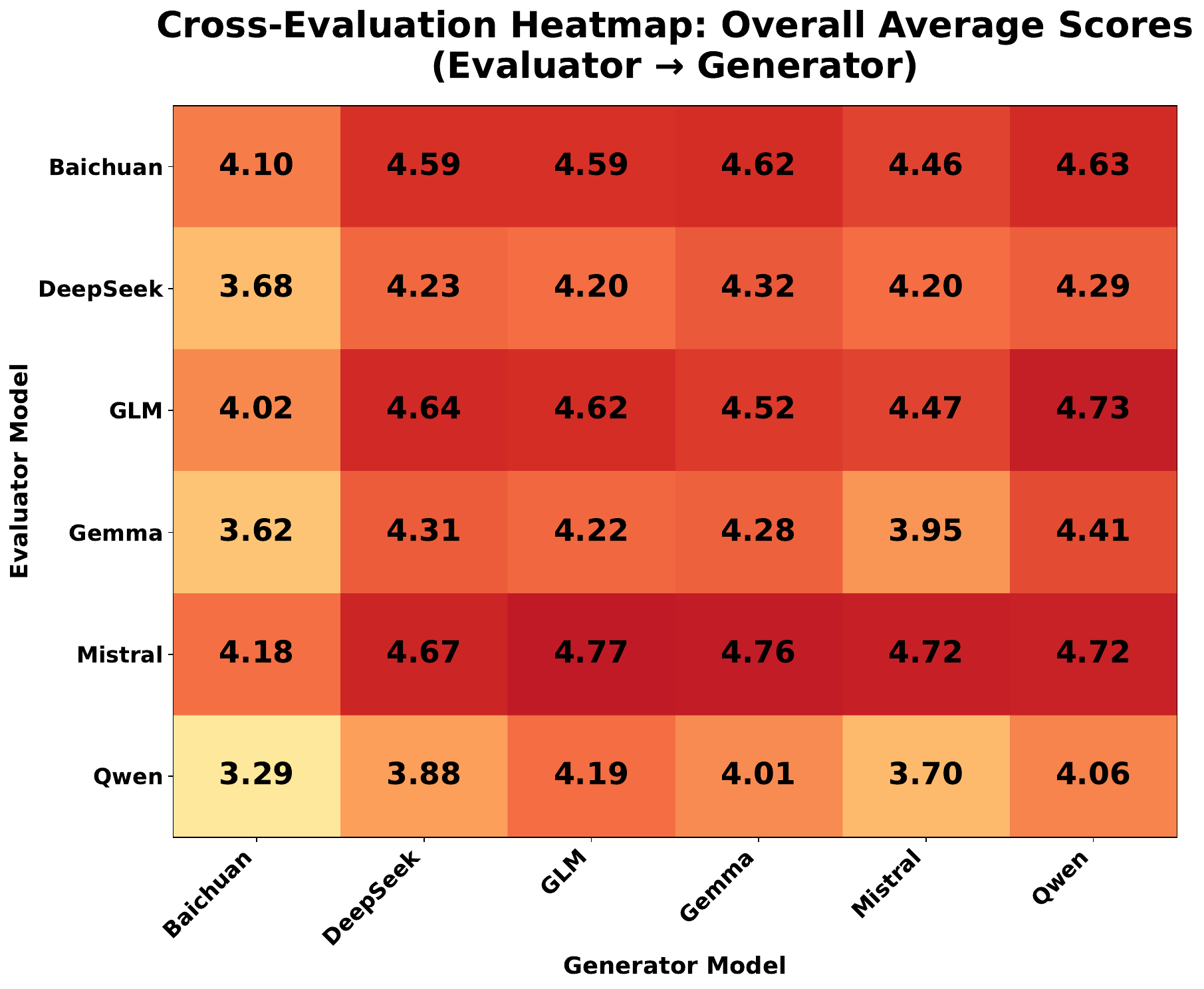}
    \caption{LLM-as-a-judge cross evaluation results. The x aches shows the generation models, while the y aches shows the evaluation models.}
    \label{fig:cross_eval_heatmap}
\end{figure}

Building upon the objective computational foundations established in analysis 1, the second analysis of our evaluation framework shifts to subjective poetry quality assessment through the LLMs themselves. This LLM-as-a-judge evaluation analysis employs the comprehensive $6 \times 6$ evaluation matrix, 
where each model serves as both generator and evaluator, creating a total of 36 combinations. 
This approach treats the collective judgment of all six models as a proxy for objective quality assessment, similar to how academic peer review aggregates expert opinions. 
To establish a baseline understanding of generation quality based on ``peer consensus'', we analyze the average scores across the five evaluation dimensions received by each generator from each evaluator (see columns in Figure \ref{fig:cross_eval_heatmap}).

\textbf{Generation Capabilities: Six models vary in generation capabilities, while the overall evaluation scores are high.} The LLM-as-a-judge evaluation heatmap shows that LLMs like Qwen, Gemma, GLM, DeepSeek as generators consistently receive higher scores across, as indicated by the predominantly red coloring, while Baichuan and Mistral receive the lower scores across evaluators, shown by the lighter, less red coloring in its row. However, we still notice that the overall evaluation scores of all models are above 3, showing the high LLM-generated consensus for the overall quality assessment across models.

\begin{figure}[!htbp]
    \centering
    \includegraphics[width=1\linewidth]{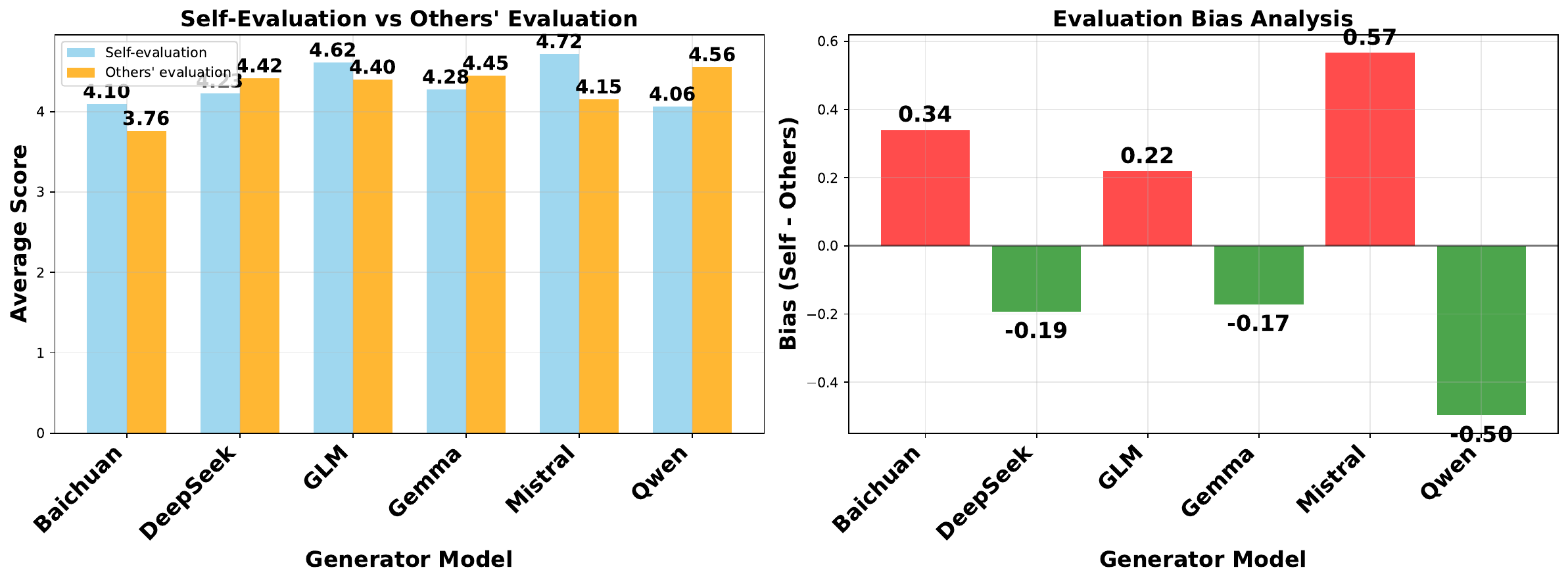}
    \caption{Bias between self and other-LLM-as-a-judge Evaluation. Left: Average scores of self vs. others' evaluation; Right: Bias (self $-$ others) results.} 
    \label{fig:bias}
\end{figure}

\textbf{Evaluator Characterization: The evaluators have their scoring tendencies and biases.} 
We then look into each evaluator model's evaluation results on different generator models (based on the horizontal view on the y aches on Figure \ref{fig:cross_eval_heatmap}). We notice that the models Mistral, GLM, Baichuan evaluate with generally higher scores across all the generator models (showing darker colors on heatmap), while DeepSeek, Gemma, Qwen give lower scores with lighter colors on the heatmap. We further show in Figure \ref{fig:bias} the evaluation results of the models for self-evaluation (i.e., the average results of the same model evaluating its generated poems) vs. other-LLM-as-judges evaluation (i.e., the average results of other four LLMs-as-judges evaluating the generated poems). We also show the biases of self $-$ others results. We notice that Baichuan, GLM, Mistral tend to rate self higher than other models, while DeepSeek, Gemma, Qwen tend to rate self lower than other models. Based on these findings, we notice that the models have their own scoring tendencies. 

\begin{figure}[!h]
\centering
\includegraphics[width=1\linewidth]{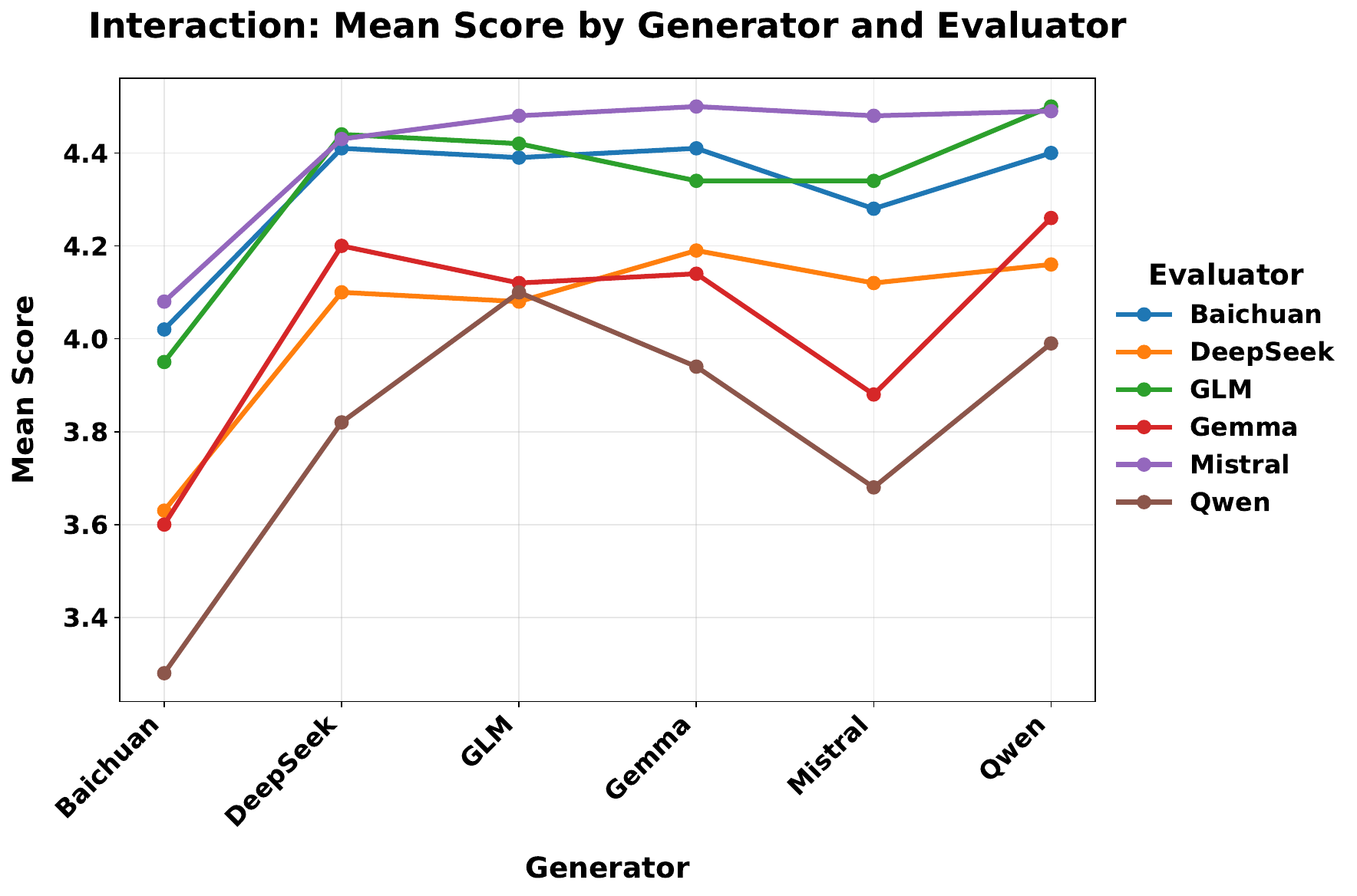}
\caption{Interaction plot of mean scores by Generator (x-axis) and Evaluator (traces).}
\label{fig:interaction-mean}
\end{figure}

\textbf{Interaction and Evidence for the ``Echo Chamber''.} 
Figure~\ref{fig:interaction-mean} plots the mean scores with generator models on the x-axis and evaluator models as separate traces.  
The pronounced non-parallel lines indicate a strong generator-evaluator interaction, the visual signature of a \textbf{methodological echo chamber}. In news and social media, an echo chamber describes an environment in which participants repeatedly encounter and reinforce the same beliefs, insulated from external challenge \cite{Cinelli2021EchoChamber}.  
We use this term to describe a closed evaluative loop among LLMs in which models, trained on overlapping data and architectures, reinforce one another's patterns rather than reflecting true poetic quality.   
This pattern shows that each evaluator adjusts its scoring behavior depending on the generator under review, suggesting collective reinforcement of shared biases rather than independent, objective judgment.

\begin{figure}[!h]
    \centering
    \includegraphics[width=1\linewidth]{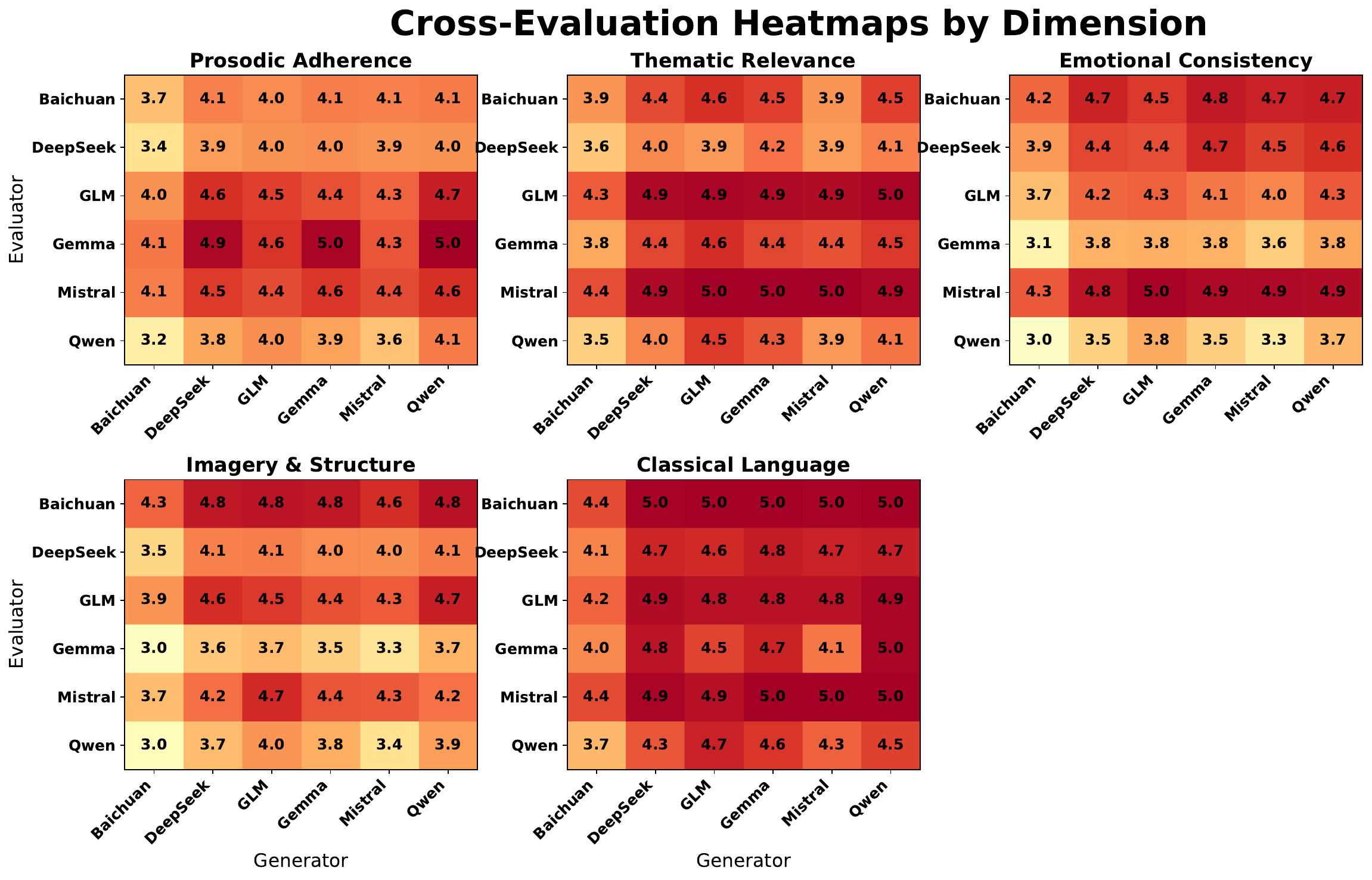}
\caption{LLM-as-a-judge evaluation scores for each dimension.}
    \label{fig:cross_eval_by_dim}
\end{figure}

\paragraph{Dimensional Analysis: Bias Patterns Across Evaluation Criteria.}
We show the detailed LLM evaluation for each dimension in Figure \ref{fig:cross_eval_by_dim}. It reveals that bias patterns are not uniform across evaluation criteria. 
The \textit{Prosodic Adherence} dimension shows the most consistent scoring patterns across evaluators, suggesting models have developed reliable criteria for formal poetry assessment. Bias variations are minimal, indicating this represents the most ``objective'' evaluation dimension. The content-focused \textit{Thematic Relevance} and \textit{Emotional Consistency} dimensions exhibit moderate bias variations, with GLM and Mistral showing particular confidence in their own thematic and emotional expression capabilities. The \textit{Imagery \& Structure} and \textit{Classical Language Authenticity} represent the most subjective dimensions, where cultural knowledge and aesthetic judgment are paramount. Here, bias patterns are most pronounced, with Qwen showing notably harsh self-assessment in Classical Language despite peer recognition of its linguistic sophistication.

\subsection{Human Evaluation}

The final analysis introduces human expert evaluation as the ultimate \emph{gold standard} for poetic quality.  
While the first two tiers revealed the internal logic and biases of LLM-based assessment, only expert judgment provides a definitive reference for both artistic and technical merit.  
A stratified random sample of 150 poems: 50 each from Qwen, DeepSeek, and Gemma, the three highest-ranked generators in Tier~2, was scored by 2 professional scholars of classical Chinese poetry using the same five-dimensional rubric applied in earlier tiers.

\textbf{Human ratings diverge sharply from machine consensus.} 
Qwen was judged a clear winner, achieving a mean score of 4.53.  
Gemma (mean 3.76) and DeepSeek (3.72) lagged far behind, despite their near-parity with Qwen in the automated evaluation.  
Dimensional analysis explains the discrepancy.  
Qwen received uniformly high marks, especially in \textit{Imagery \& Structure} (4.96) and \textit{Prosodic Adherence} (4.46).  
Gemma, although strong in \textit{Thematic Relevance} (4.56), failed catastrophically on \textit{Prosodic Adherence} (mean 1.72) due to frequent rule violations and occasional English intrusions in Chinese verses, while this has not been noticed by LLM evaluations.  
DeepSeek showed moderate competence across dimensions but lacked both technical precision and artistic flair.  
These findings highlight limitations of machine  consensus, which missed Gemma's critical prosodic flaws. See the detailed results and visualization of human evaluation scores in Appendix \ref{sec:additional results human}.

\textbf{Statistical alignment of human and LLM scores.}
We computed Spearman correlations between human and machine scores across all 150 poems (Table~\ref{tab:spearman_correlation}).
At the aggregate level, modest but significant positive correlations emerge for \textit{Thematic Relevance} ($\rho=0.309$, $p<0.001$) and \textit{Language Authenticity} ($\rho=0.297$, $p<0.001$), indicating a shared but limited sense of quality.

Generator-level analysis reveals sharper divergences.
For Qwen, the top human-rated model, no dimension showed significant correlation, demonstrating that machine evaluators failed to capture qualities valued by experts.
Gemma provides clear evidence of systematic misalignment: a strong positive correlation on \textit{Thematic Relevance} ($\rho=0.622$, $p<0.001$) but a negative, non-significant correlation on \textit{Prosodic Adherence} ($\rho=-0.176$).
This shows that LLM evaluators collectively overlooked Gemma's 
prosodic violations.

\begin{table}[!h]
\centering
\scriptsize
\setlength\tabcolsep{3pt}
\renewcommand\arraystretch{0.85}
\begin{tabular}{@{}lccccc@{}}
\toprule
\textbf{Generator} & \textbf{Prosody} & \textbf{Thematic} & \textbf{Emotional} & \textbf{Imagery} & \textbf{Language} \\
\midrule
Overall (N=150) & 0.196* & 0.309*** & 0.226** & 0.095 & 0.297*** \\
Qwen (n=50)     & -0.033 & 0.225    & 0.017   & -0.161 & 0.052 \\
DeepSeek (n=50) & 0.097  & 0.152    & 0.354** & 0.038  & 0.225 \\
Gemma  (n=50)   & -0.176 & 0.622*** & 0.304*  & 0.113  & 0.604*** \\
\bottomrule
\multicolumn{6}{l}{\tiny Significance: * $p<0.05$, ** $p<0.01$, *** $p<0.001$}
\end{tabular}
\caption{Spearman correlations ($\rho$) between human and LLM judges scores. Detailed visualization of the correlations is shown in Appendix \ref{sec:additional results human}.}
\label{tab:spearman_correlation}
\end{table}

\textbf{Validating the ``Echo Chamber'' hypothesis.} We empirically validate this hypothesis through Gemma's specific failure case. While human judges identified significant prosodic defects, the LLM evaluator cohort failed to penalize them, demonstrating a collective reinforcement of errors. This phenomenon is rigorously quantified by the Spearman Rank Correlation ($\rho$) in Table \ref{tab:spearman_correlation}. A striking example is the negative correlation for Gemma's Prosodic Adherence ($\rho=-0.176$), which stands in sharp contrast to the high alignment observed in semantic tasks like Thematic Relevance ($\rho=0.622, p < 0.001$). This disparity suggests that while LLMs align on semantic content, they share a critical blind spot regarding structural constraints. We attribute this to overlapping limitations in pretraining data, i.e., the presence of cross-lingual artifacts and a lack of rigorous, rule-based prosodic representation across the models.

\section{Discussion}

Our findings show that LLMs can produce surface-level fluency in classical Chinese poetry while exhibiting fundamental weaknesses that automated metrics fail to detect.  Using a three-step evaluation framework including computational analysis, cross-model scoring, and expert review, we reveal 
model capabilities in literacy, 
evaluation biases, and cultural competence, 
with following insights:

\textbf{The most striking result is an echo-chamber effect: models trained on similar data converge on shared yet inaccurate standards for poetic quality.}  
Gemma's severe prosodic failures received ratings nearly equivalent to Qwen's technically correct output when judged by other LLMs, showing that cross-model consensus does not guarantee accuracy.  
This undermines the common assumption (e.g., \citealp[]{liu-etal-2023-g,wang-etal-2024-large-language-models-fair,chan2024chateval,hu2025languagemodelpreferenceevaluation}) that ensemble evaluation reduces bias; 
semantic alignment (>0.94) suggests that common training data instead amplifies collective blind spots.  
In creative 
tasks, human oversight therefore remains indispensable.

\textbf{Single metrics such as entropy cannot capture the multifaceted nature of poetic quality.} 
Entropy and other one-dimensional scores provide useful diagnostics but cannot serve as stand-alone measures.  
The ``entropy paradox'', where low entropy reflects either strong formal control or restricted vocabulary, shows how isolated metrics obscure key distinctions.  
Combining entropy with single ``fingerprints'' and semantic-coherence measures proved more informative, revealing, for example, Baichuan's template-driven generation.  
Future work should adopt multi-metric diagnostic frameworks \cite{ghazvininejad-etal-2017-hafez,van-de-cruys-2020-automatic} rather than simple performance rankings.

\textbf{High semantic similarity masks a lack of cultural and stylistic depth.}  
Despite striking inter-model agreement, systems failed to capture core cultural associations or to differentiate canonical poet styles.  
The inability to represent the classic ``willow-sadness'' motif exemplifies this gap.  
Such results temper optimistic claims (e.g., \citealp[]{deng2024turing,Yu2024,qu2025poetoneframeworkconstrainedgeneration}) that LLM poetry approaches human quality: the alignment paradox shows that models may agree semantically on styles they cannot actually realize.

\textbf{Models recombine learned patterns rather than create genuinely new poetry.}  
High metric overlap of traditional imagery and frequent reuse of canonical lines indicate recombination rather than genuine innovation \cite{10.1145/3442188.3445922}. LLM output largely falls into the so-called ``combinatorial creativity'' - producing novel combinations of familiar ideas \cite{boden2003creative}, as in humans.
Even Qwen's superior performance reflects precise pattern matching and constraint satisfaction more than novel literary creation, reinforcing critiques of LLMs as advanced auto-completion systems.

\textbf{A three-step evaluation framework offers a path toward more reliable creative-AI assessment.}  
Our framework-integrating computational diagnostics, cross-model evaluation, and expert judgment-provides a transferable methodology for assessing creative AI in culturally rich domains.  
Future research should (i) design metrics that jointly capture structural fidelity and cultural nuance, (ii) diversify training and evaluation data to reduce shared blind spots, and (iii) emphasize explicit structural constraints over vague stylistic prompts.

\textbf{Qualitative analysis.} Appendix \ref{sec:qualitative} further presents a few qualitative cases of the LLM generated poems.

\section{Conclusion}

Our study demonstrates that while state-of-the-art LLMs can generate fluent Tang-style poetry, they exhibit critical limitations that surface-level metrics and cross-model evaluation often fail to detect. 
The proposed three-step framework, combining computational analysis, LLM-based assessment, and human expert validation, reveals systematic biases, including echo-chamber effects, shared poem outputs  and shallow cultural association fidelity. Models tend to recombine learned patterns rather than create genuinely novel poetic expressions, and high inter-model agreement does not guarantee alignment with human judgments.  

These findings underscore the continued importance of human oversight in culturally and structurally complex creative tasks. Future work should focus on multi-metric diagnostic evaluation, integration of explicit structural and cultural constraints, and strategies to mitigate shared blind spots in LLM assessment. Overall, our results highlight both the potential and the current limitations of LLMs in creative language generation, providing a roadmap for more reliable and culturally informed AI evaluation and development.

\section*{Limitations}

The focus on classical Chinese poetry, while providing cultural specificity and technical constraints necessary for rigorous evaluation, could have the limits the generalizability of findings to other creative domains in different languages. Future research should extend this methodological approach to other languages in digital literacy domains to test the multilingual and multicultural capabilities of the models, thereby 
to test the broader applicability of the echo chamber effect and evaluation framework.

The computational metrics we employed, particularly semantic similarity measurements, were limited by their reliance on models trained primarily on prosaic text. Future work should develop evaluation metrics specifically designed for poetic and metaphorical language, potentially incorporating specialized embeddings trained on literary corpora or rule-based systems that can capture formal poetic structures.

A further limitation concerns the study's scope regarding prompt sensitivity and latent content biases. We did not exhaustively test diverse prompting architectures, such as Chain-of-Thought \cite{wei2022chain} or role-playing protocols, which might enhance poetic reasoning. Additionally, the models' underlying safety alignments may suppress certain historical themes or archaic expressions. Future work should specifically investigate how to decouple these biases from valid creative constraints to improve generation authenticity.

\section*{Ethical Considerations}

We address some potential ethical considerations regarding the paper in this section.

\paragraph{Risk Concerns.} This work examines LLMs in the culturally sensitive domain of Tang poetry generation. While the generated texts draw on public-domain traditions, they may misrepresent or oversimplify cultural values if interpreted outside research contexts. We caution against deploying LLM-generated poems as authentic cultural artifacts and emphasize that they should only serve as analytical probes of model capabilities. Our study also reveals systematic evaluation biases, underscoring the risks of over-reliance on automated judgments without human validation. By releasing code and generated outputs for research purposes only (the code and the outputs have been submitted together with the paper as supplementary documents and will be publicly released upon paper publication), we aim to promote transparency while minimizing risks of misappropriation. 

\paragraph{Use of AI Assistants.} The authors acknowledge the use of ChatGPT solely for correcting grammatical errors, enhancing the coherence of the final manuscript, and providing assistance with coding.

\bibliography{custom}

\appendix
\label{sec:appendix}

\section{Poetry Dimensions for Generation}
\label{sec:dim_zh}
We cover the five important dimensions in Tang poetry: poetic form (\begin{CJK}{UTF8}{gbsn}格式\end{CJK}), target poet style (\begin{CJK}{UTF8}{gbsn}诗人\end{CJK}), theme (\begin{CJK}{UTF8}{gbsn}主题\end{CJK}), emotion (\begin{CJK}{UTF8}{gbsn}情感\end{CJK}), and imagery (\begin{CJK}{UTF8}{gbsn}意象\end{CJK}), with each including representative specific elements (Table~\ref{tab:poetry_dimensions_zh}).

\begin{table}[htbp]
\centering
\small
\begin{tabular}{|l|l|}
\hline
\textbf{Dim.}& \textbf{Specific Elements} \\
\hline
\begin{CJK}{UTF8}{gbsn}格式\end{CJK} & \begin{CJK}{UTF8}{gbsn}五/七言绝句、五/七言律诗\end{CJK} \\
\hline
\begin{CJK}{UTF8}{gbsn}诗人\end{CJK} & \begin{CJK}{UTF8}{gbsn}李白、杜甫、白居易、王维、李商隐\end{CJK} \\
\hline
\begin{CJK}{UTF8}{gbsn}主题 \end{CJK}& \begin{CJK}{UTF8}{gbsn}山水、思乡、怀古、田园、送别\end{CJK} \\
\hline
\begin{CJK}{UTF8}{gbsn}情感\end{CJK} & \begin{CJK}{UTF8}{gbsn}悲伤、宁静、豪放、浪漫、喜悦 \end{CJK}\\
\hline
\begin{CJK}{UTF8}{gbsn}意象\end{CJK} & \begin{CJK}{UTF8}{gbsn}风、花、柳、月、雁\end{CJK} \\
\hline
\end{tabular}
\caption{Poetry dimensions and elements for generation in Chinese. The English translation is in \S\ref{sec:generation}.}
\label{tab:poetry_dimensions_zh}
\end{table}

\section{Generation and Evaluation Prompt Templates}
\label{sec:prompt}
\paragraph{Generation prompts.}  
For each model, the generation prompts provided explicit instructions regarding the required poetic form (e.g., ``five-character quatrain''), the style of a specific poet (e.g., ``in the style of Li Bai''), the thematic content (e.g., ``farewell''), the emotion (e.g., ``melancholy''), and the imaginary (e.g., ``willows''). An example prompt (with translation in English) is shown below:

\begin{quote}
\textbf{ZH:}
\begin{CJK}{UTF8}{gbsn}
请模仿\underline{杜甫}的风格，创作一首关于\underline{送别}的\underline{七言绝句}，体现\underline{悲伤}情感，并包含\underline{柳}的意象。请只输出诗歌本身。
\end{CJK}
\end{quote}

\begin{quote}
\textbf{EN:}
\texttt{Please imitate the style of \underline{Du Fu} and compose a \underline{seven-character quatrain} about \underline{farewell}, conveying \underline{melancholy} and including the imagery of \underline{willows}. Output only the poem itself.}
\end{quote}

\paragraph{Evaluation prompts.}  
Evaluation prompts instructed the model to assess a given poem on five dimensions: \textit{Prosodic Adherence}, \textit{Thematic Relevance}, \textit{Emotional Consistency}, \textit{Imagery \& Structure}, and \textit{Classical Language Authenticity}. Each dimension required an integer score from 1 to 5, returned in structured JSON. The original Chinese prompt and its translation are shown as follows:

\begin{quote}
\begin{CJK}{UTF8}{gbsn}
\textbf{ZH:}
\begin{verbatim}
# 角色
你是一位中国古典文学评价专家。请对以下诗歌进行评分，每个维度给出1-5的整数分数。

# 评分标准
1分 = 很差
2分 = 较差  
3分 = 一般
4分 = 良好
5分 = 优秀

# 评价维度
1. **韵律格律 (prosodic adherence)**: 音韵、平仄、押韵、节奏
2. **主题切合度 (thematic relevance)**: 主题明确性与契合度
3. **情感一致性 (emotional consistency)**: 情感表达的真挚性与一致性
4. **意象与结构 (imagery structure)**: 意象生动性与结构合理性
5. **语言经典性 (language authenticity)**: 语言的古典韵味与准确性

# 诗歌信息
- **诗人**: {poet}
- **主题**: {theme}
- **情感**: {emotion}
- **意象**: {imagery}
- **形式**: {form}
- **诗歌正文**: {poem_text}

# 输出要求
请直接输出JSON格式的评分结果，每个维度只需给出1-5的整数分数：

{
  "prosodic_adherence": 分数,
  "thematic_relevance": 分数,
  "emotional_consistency": 分数,
  "imagery_structure": 分数,
  "language_authenticity": 分数
}
\end{verbatim}
\end{CJK}
\end{quote}

\begin{quote}
\textbf{EN:}
\begin{myverb}
# Role
You are an expert in evaluating classical Chinese literature. Please score the following poem, giving an integer from 1 to 5 for each dimension.

# Scoring Criteria
1 = Very Poor  
2 = Poor  
3 = Fair  
4 = Good  
5 = Excellent

# Evaluation Dimensions
1. **prosodic adherence**: Rhyme, tonal pattern, rhyme scheme, rhythm  
2. **thematic relevance**: Clarity and appropriateness of theme  
3. **emotional consistency**: Sincerity and consistency of emotional expression  
4. **imagery structure**: Vividness of imagery and soundness of structure  
5. **language authenticity**: Classical flavor and accuracy of language

# Poem Information
- **Poet**: {poet}  
- **Theme**: {theme}  
- **Emotion**: {emotion}  
- **Imagery**: {imagery}  
- **Form**: {form}  
- **Poem Text**: {poem_text}

# Output Requirement
Please output the scoring results directly in JSON format, giving an integer 1-5 for each dimension:

{
  "prosodic_adherence": score,
  "thematic_relevance": score,
  "emotional_consistency": score,
  "imagery_structure": score,
  "language_authenticity": score
}
\end{myverb}
\end{quote}

\section{Poem Generation Procedure}
\label{sec:poem generation}
\begin{figure}[ht]
    \centering
    \includegraphics[width=1\linewidth]{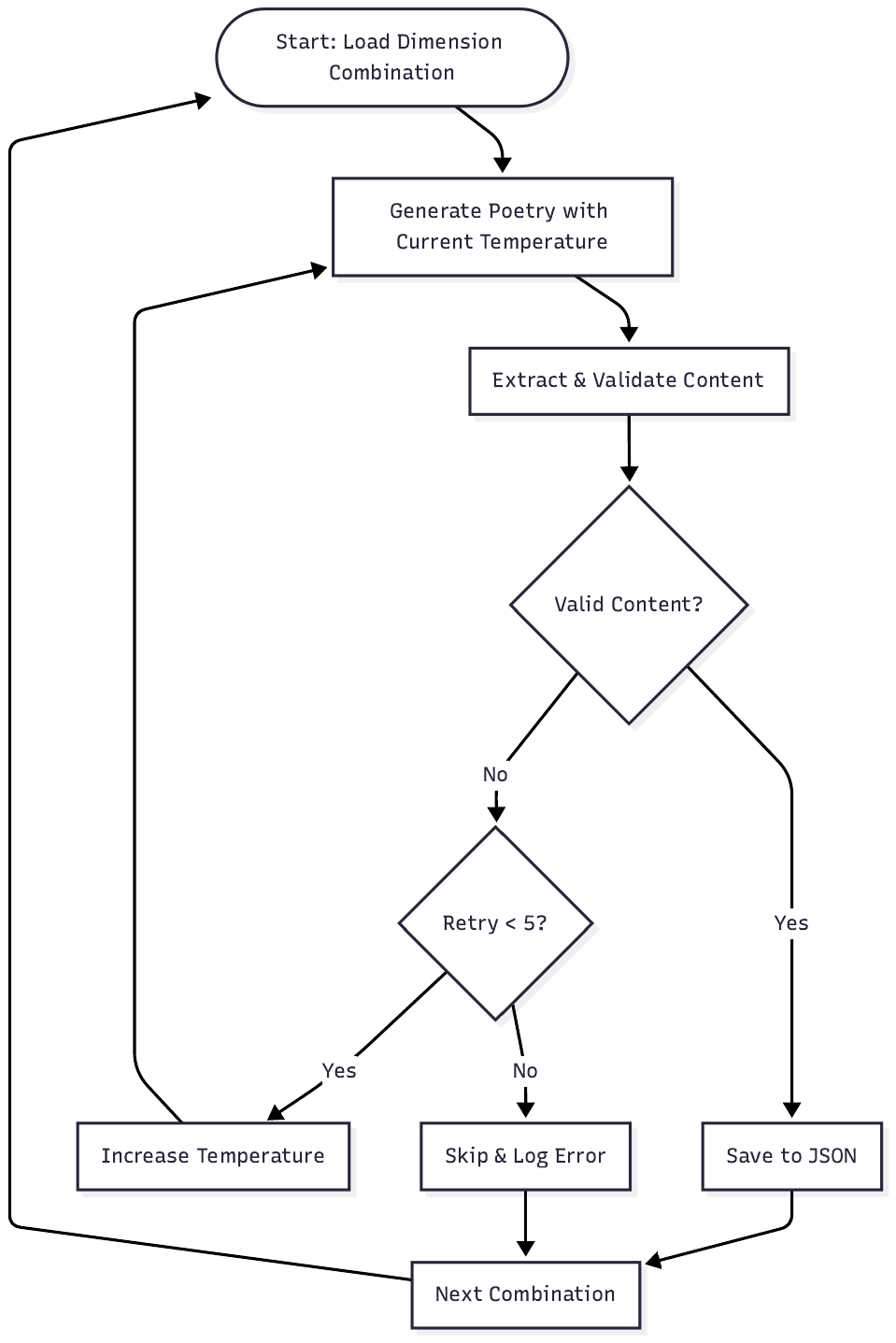}
    \caption{Poem Generation Flow Chart. The experimental loop generated poems for combinations of all dimension elements, after which outputs were cleansed and validated against structural and content criteria. Failed generations triggered adaptive retries with increased temperature, and all results were stored incrementally in JSONL to ensure reliability.}
    \label{fig:poem_gen_flow_chart}
\end{figure}

For each of the six models, we independently generated poems for every combination of target poet, theme, emotion, imagery, and form. Prompts were derived from a fixed template shown in the previous section (Appendix \ref{sec:prompt}) and explicitly instructed the model to satisfy all specified constraints while emulating the designated poet's style (Figure \ref{fig:poem_gen_flow_chart}).

The generation loop enumerated all task combinations using Python's \texttt{itertools.product}. For each task, the model produced a poem whose raw output was immediately cleaned to isolate the poetic text.\footnote{This initial cleaning was a lightweight check to remove empty or malformed responses and does not duplicate the subsequent validation step.} The output was then validated against basic structural and content requirements. If a poem failed validation, the system automatically retried generation up to five times, incrementally increasing the sampling temperature to promote diversity. All valid outputs were written incrementally to JSON Lines (\texttt{.jsonl}) files to prevent data loss in the event of interruption.

\section{Generated Poem Cleaning Heuristics}
Raw outputs often contain prefixes, commentary, or formatting noise. Cleaning follows a two-step procedure: first, a model-driven cleaner (e.g., Qwen2.5) is prompted to extract the text into JSON-formatted poems; if this fails, regex and heuristic rules are applied as fallback to remove explanatory phrases or extraneous lines, while retaining necessary generation traces for later analysis. The cleaned results are stored in a standardized structure for automated evaluation.

In cases where model-driven cleaning (Qwen2.5 extractor) failed to return a valid JSON object, regex-based heuristics were applied:
\begin{itemize}[leftmargin=*,noitemsep]
  \item Remove leading/trailing commentary or role markers (e.g., ``As an AI model, here is your poem:'').  
  \item Collapse multiple empty lines into a single line break.  
  \item Retain non-standard artifacts (e.g., repeated punctuation) when directly embedded in poem lines.  
\end{itemize}

The following prompt (original prompt in Chinese and translated prompt in English) was used for poem cleaning:

\begin{quote}
\begin{CJK}{UTF8}{gbsn}
\textbf{ZH:}
\begin{verbatim}
"""请从以下文本中提取纯净的诗歌内容。

**规则**：
- 保留诗歌的所有诗句和原有的换行分段
- 删除任何前缀后缀（如"好的，这是一首..."、"希望您喜欢"等）
- 删除诗歌标题（如《送别》）
- 删除任何注释、解释或旁白
- 删除与诗歌无关的对话性文字

**输出格式**：请严格按照JSON格式返回，如果没有诗歌则extracted_poem为空字符串：
{{"status": "success", "extracted_poem": 
"提取的纯净诗歌文本"}}

**待处理文本**：
{raw_text}

**回答**："""
\end{verbatim}
\end{CJK}
\end{quote}

\begin{quote}

\textbf{EN:}
\begin{myverb}

"Please extract the pure poetic content from the following text.

Rules:

- Retain all lines of the poem and the original line breaks

- Remove any prefixes or suffixes (e.g., 'This is a poem' or 'I hope you enjoy it')

- Remove the poem title (e.g., 'Farewell')

- Remove any annotations, explanations, or commentary

- Remove any dialogue-related text unrelated to the poem

Output Format: Please return the output strictly in JSON format. 
If no poem is extracted, set 'extracted_poem' to an empty string:  
{'status': 'success', 'extracted_poem': 'Extracted pure poetic text'}  

Raw Text: {raw_text}  

Response: "
%\end{verbatim}
\end{myverb}
\end{quote}

\begin{figure}[ht]
    \centering
    \includegraphics[width=1\linewidth]{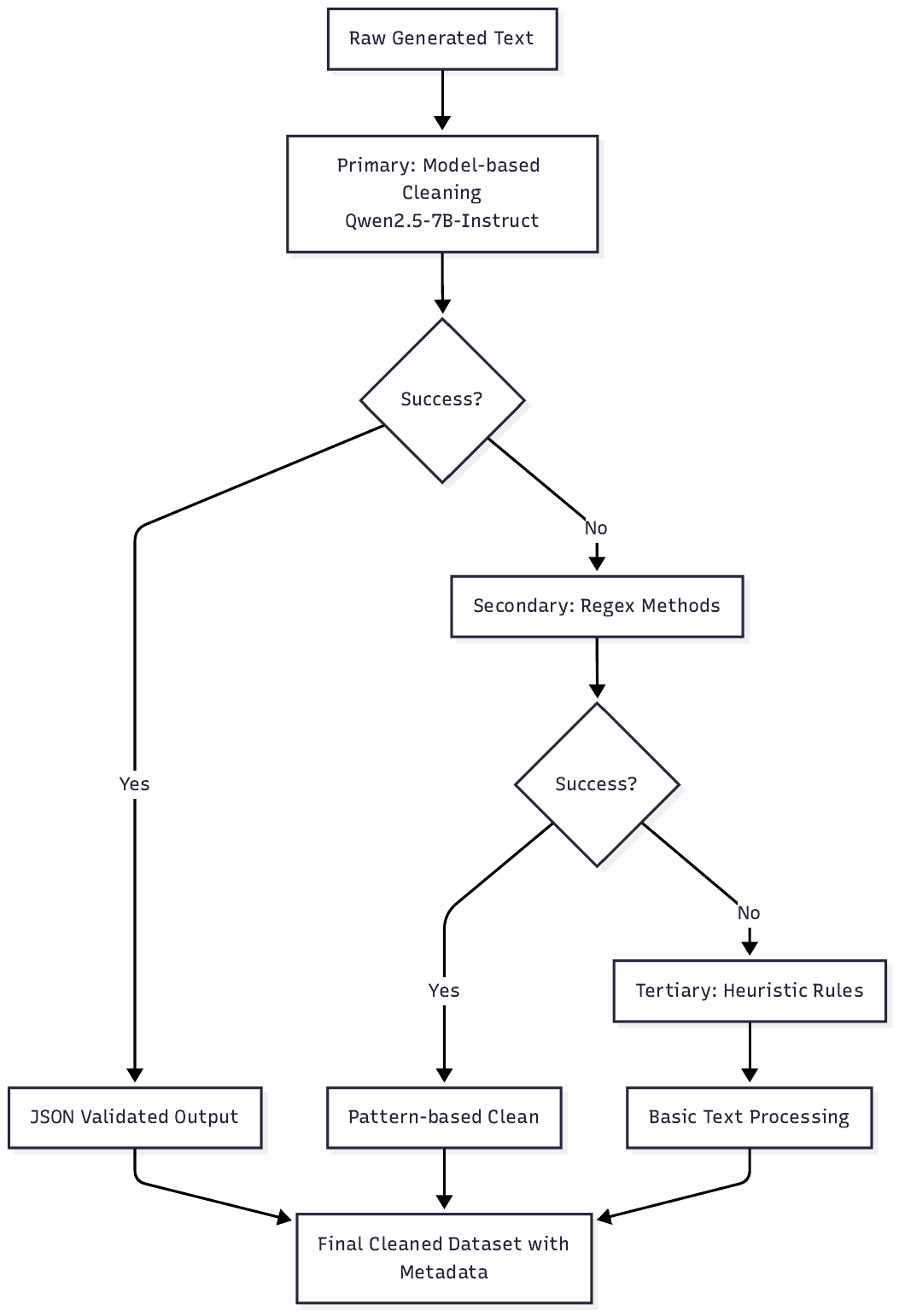}
    \caption{Poem Cleaning Flow Chart. For text cleaning, I used Qwen, which had an excellent performance for generating complete and correctly formatted poems in the initial format consistency review. A structured prompt guided the model to remove extraneous content and output the cleaned text strictly as a JSON object.}
    \label{fig:poem_cleaning}
\end{figure}

The flow for data cleaning could be seen in Figure \ref{fig:poem_cleaning}

\section{Embedding models}

Sentence-level embeddings were obtained using a Chinese sentence transformer (compatible with \citet{reimers2019sentence}). For coherence analysis, cosine similarity was computed both across poem-prompt pairs and between adjacent lines.

\section{Hyperparameters}

\begin{itemize}[leftmargin=*,noitemsep]
  \item \textbf{Generation temperature:} $T=0.4$ (with up to 2 retries for diversity).  
  \item \textbf{Evaluation temperature:} $T=0.2$ (to enforce deterministic scoring).  
  \item \textbf{Maximum length:} 128 tokens (covering the longest regulated verse).  
\end{itemize}

\section{Data Formats and Storage}

All cleaned outputs were stored in structured JSON with metadata including:
\begin{itemize}[leftmargin=*,noitemsep]
  \item Model identifier  
  \item Prompt specification (form, style, theme, emotion, imagery)  
  \item Generated poem text  
  \item Evaluation scores (when available)  
\end{itemize}
This design ensured automatic parsing for downstream statistical analysis.

\section{Statistical Procedures}

Entropy comparisons were conducted using one-way ANOVA, with Tukey HSD for post-hoc tests. For evaluator-generator interactions, linear mixed-effects models were implemented in \texttt{lme4} \cite{Bates2015} (R package), treating generator/evaluator as fixed and random effects respectively.

\section{Human Expert Evaluation}

One of the leading authors and another project collaborator were asked to conduct the evaluation of the 150 samples (500 each from Qwen, DeepSeek, and Gemma, the three highest-ranked generators 
in Step 2 LLM evaluations) as volunteers. There exists disagreement on minor cases that were resolved after discussion. Both annotators are consent about the annotated data use.

\section{Qualitative Analysis}
\label{sec:qualitative}

While large-scale metrics provide an essential overview of model performance, close reading of individual outputs exposes error patterns that aggregate statistics cannot capture.  
This section examines 150 poems (50 each) generated by the three top-performing LLMs, Qwen, Gemma, and DeepSeek, highlighting characteristic strengths and weaknesses.

\textbf{DeepSeek: Structural Confusion and Collage Effects.}  
DeepSeek frequently produced text that was descriptive commentary rather than original poetry.  
For example, when prompted for a five-character quatrain on ``homesickness'' in the style of Li Bai, it generated a prose-like analysis of Li Bai's famous poem (Table~\ref{tab:model-output-example-translated}).  
Other outputs resembled collages of celebrated lines from multiple poets (e.g., Gao Shi, Cen Shen, Liu Zongyuan, and even Mao Zedong), resulting in structurally incoherent verse without a unified voice or theme.  
Nevertheless, DeepSeek generally respected basic distinctions among quatrains, regulated verse, and line length, suggesting only a partial grasp of formal constraints.

\begin{table}[!htbp]
\centering
\scriptsize
\setlength\tabcolsep{1pt}
\begin{tabular}{lll}
\toprule
\textbf{Dim.} & \textbf{Elements} & \textbf{English Translation} \\ 
\midrule
Poet    & \begin{CJK}{UTF8}{gbsn}李白\end{CJK}     & Li Bai \\
Theme   & \begin{CJK}{UTF8}{gbsn}思乡\end{CJK}     & Homesickness \\
Emotion & \begin{CJK}{UTF8}{gbsn}悲伤\end{CJK}     & Sadness \\
Imagery & \begin{CJK}{UTF8}{gbsn}月亮\end{CJK}       & Moon \\
Form    & \begin{CJK}{UTF8}{gbsn}五言绝句\end{CJK} & Five-character Quatrain \\ 
\midrule
\multicolumn{3}{c}{\textbf{Generated Text (Original)}} \\ 
\midrule
\multicolumn{3}{p{0.48\textwidth}}{%
\begin{CJK}{UTF8}{gbsn}
是唐代诗人李白的代表作之一，以简洁明快的语言和深远的意境，表达了作者对故乡的深深思念。诗中``床前明月光''描绘了一个宁静的夜晚，月光如水，洒在床前，营造出一种幽远而神秘的氛围。后两句``疑是地上霜''则巧妙地运用比喻，将月光比作秋霜，既展现了月光的皎洁，又暗示了季节的更替，引发读者对时光流逝的感慨。整首诗...
\end{CJK}
} \\ 
\midrule
\multicolumn{3}{c}{\textbf{Generated Text (Translation)}} \\ 
\midrule
\multicolumn{3}{p{0.48\textwidth}}{%
This is one of the representative works of the Tang Dynasty poet Li Bai, which uses concise language and profound artistic conception to express the author's deep longing for his hometown. In the poem, 'Bright moonlight before my bed' depicts a quiet night where the moonlight, like water, spills before the bed, creating a distant and mysterious atmosphere. The latter two lines, 'I mistake it for frost on the ground,' cleverly use a metaphor, comparing the moonlight to autumn frost, which not only shows the brightness of the moon but also hints at the changing of seasons, evoking the reader's sigh over the passage of time. The entire poem...
} \\ 
\bottomrule
\end{tabular}
\caption{Example of a descriptive model output with English translation.}
\label{tab:model-output-example-translated}
\end{table}

\textbf{Gemma: Systematic Format Violations and Cross-Lingual Intrusions.}  
Gemma's strongest semantic performance was undermined by persistent failures of form.  
It routinely defaulted to seven-character quatrains, even when explicitly instructed to produce other forms such as five-character regulated verse.  
A second recurring error was cross-lingual contamination, with English words embedded within Chinese lines (Table~\ref{tab:model-error-example}).  
Table~\ref{tab:model-format-error} illustrates another case where Gemma ignored the requested five-character regulated structure.  
Despite these technical lapses, human evaluators often rated Gemma highly for theme, emotion, and imagery, underscoring a notable gap between semantic quality and formal correctness.

\begin{table}[htbp]
\centering
\scriptsize
\setlength\tabcolsep{1pt}
\begin{tabular}{@{}lll@{}}
\toprule
\textbf{Dim.} & \textbf{Elements} & \textbf{English Translation} \\ 
\midrule
Poet & \begin{CJK}{UTF8}{gbsn}李白\end{CJK} & Li Bai \\
Theme & \begin{CJK}{UTF8}{gbsn}怀古\end{CJK} & Nostalgic \\
Emotion & \begin{CJK}{UTF8}{gbsn}豪放\end{CJK} & Bold \\
Imagery & \begin{CJK}{UTF8}{gbsn}花\end{CJK} & Flower \\
Form & \begin{CJK}{UTF8}{gbsn}七言绝句\end{CJK} & Seven-character Quatrain \\ \midrule
\multicolumn{3}{c}{\textbf{Generated Poem}} \\
\midrule
\textbf{Original Chinese} & \multicolumn{2}{l}{\textbf{English Translation}} \\ \cmidrule(l){1-1} \cmidrule(l){2-3}
\begin{CJK}{UTF8}{gbsn}落花飘零满地红，\end{CJK} & \multicolumn{2}{l}{Fallen flowers drift down, covering the ground in red,} \\
\begin{CJK}{UTF8}{gbsn}旧梦如烟随风blown。\end{CJK} & \multicolumn{2}{l}{Old dreams, like smoke, are blown with the wind.} \\
\begin{CJK}{UTF8}{gbsn}举杯邀月寄相思，\end{CJK} & \multicolumn{2}{l}{I raise my cup to the moon, sending my longing,} \\
\begin{CJK}{UTF8}{gbsn}何处寻得当年雄。\end{CJK} & \multicolumn{2}{l}{Where can I find the hero of bygone years?} \\ \bottomrule
\end{tabular}
\caption{Example of a Model Output with Cross-Lingual Contamination.}
\label{tab:model-error-example}
\end{table}


\begin{table}[htbp]
\centering
\scriptsize
\setlength\tabcolsep{1pt}
\begin{tabular}{@{}lll@{}}
\toprule
\textbf{Dim.} & \textbf{Elements} & \textbf{English Translation} \\ \midrule
Poet & \begin{CJK}{UTF8}{gbsn}白居易\end{CJK} & Bai Juyi \\
Theme & \begin{CJK}{UTF8}{gbsn}怀古\end{CJK} & Nostalgic \\
Emotion & \begin{CJK}{UTF8}{gbsn}豪放\end{CJK} & Bold \\
Imagery & \begin{CJK}{UTF8}{gbsn}月\end{CJK} & Moon \\
\textbf{Requested Form} & \textbf{\begin{CJK}{UTF8}{gbsn}五言律诗\end{CJK}} & \textbf{Five-character Regulated Verse} \\
\textbf{Actual Form} & \textbf{\begin{CJK}{UTF8}{gbsn}七言绝句\end{CJK}} & \textbf{Seven-character Quatrain} \\ \midrule
\multicolumn{3}{c}{\textbf{Generated Poem}} \\
\midrule
\textbf{Original Chinese} & \multicolumn{2}{l}{\textbf{English Translation}} \\ \cmidrule(l){1-1} \cmidrule(l){2-3}
\begin{CJK}{UTF8}{gbsn}旧梦如烟消残阳，\end{CJK} & \multicolumn{2}{l}{Old dreams, like smoke, fade in the setting sun,} \\
\begin{CJK}{UTF8}{gbsn}明月照破落城墙。\end{CJK} & \multicolumn{2}{l}{The bright moon shines on the dilapidated city walls.} \\
\begin{CJK}{UTF8}{gbsn}英雄泪湿青史册，\end{CJK} & \multicolumn{2}{l}{A hero's tears wet the pages of history,} \\
\begin{CJK}{UTF8}{gbsn}多少悲欢今何在？\end{CJK} & \multicolumn{2}{l}{How many joys and sorrows, where are they now?} \\ \bottomrule
\end{tabular}
\caption{Example of a Model Output Exhibiting a Format Violation.}
\label{tab:model-format-error}
\end{table}

\textbf{Qwen: Consistent Formal Control with Minor Literary Issues.}  
Qwen displayed near-perfect adherence to prescribed forms and meter.  
Format violations were rare, and outputs consistently satisfied structural and stylistic requirements.  
Remaining issues were subtle: occasional thematic ambiguity or slight rhythmic unevenness that affected aesthetic polish but not correctness.  
This reliability confirms Qwen as the most technically proficient generator in the study.

\section{Additional Results -- Information Gain}
\label{sec:additional results information gain}
To quantify how different creative dimensions in the prompts influence the diversity of the generated poems, we analyzed the information gain for each dimension, drawing from \citet{ma-etal-2025-algorithmic}.  
It measures how much information one random variable provides about another. It is calculated as the difference between the entropy of the variable from general group (Group) and its conditional entropy given another variable from a specific subgroup (Subgroup):

\begin{equation}
I(X; Y) = H(Y) - H(Y \mid X)
\end{equation}

It indicates how much knowing one variable (e.g., $X$) reduces uncertainty about another variable (e.g., $Y$). 
A higher information gain indicates that knowing one variable reduces uncertainty about another variable. A significant drop in entropy indicates that a given dimension strongly constrains the model's vocabulary, forcing it into a more predictable and focused generative mode. The analysis reveals which dimensions have the most and least impact on the models' outputs.

\begin{figure*}[htbp]
    \centering
    \includegraphics[width=1\linewidth]{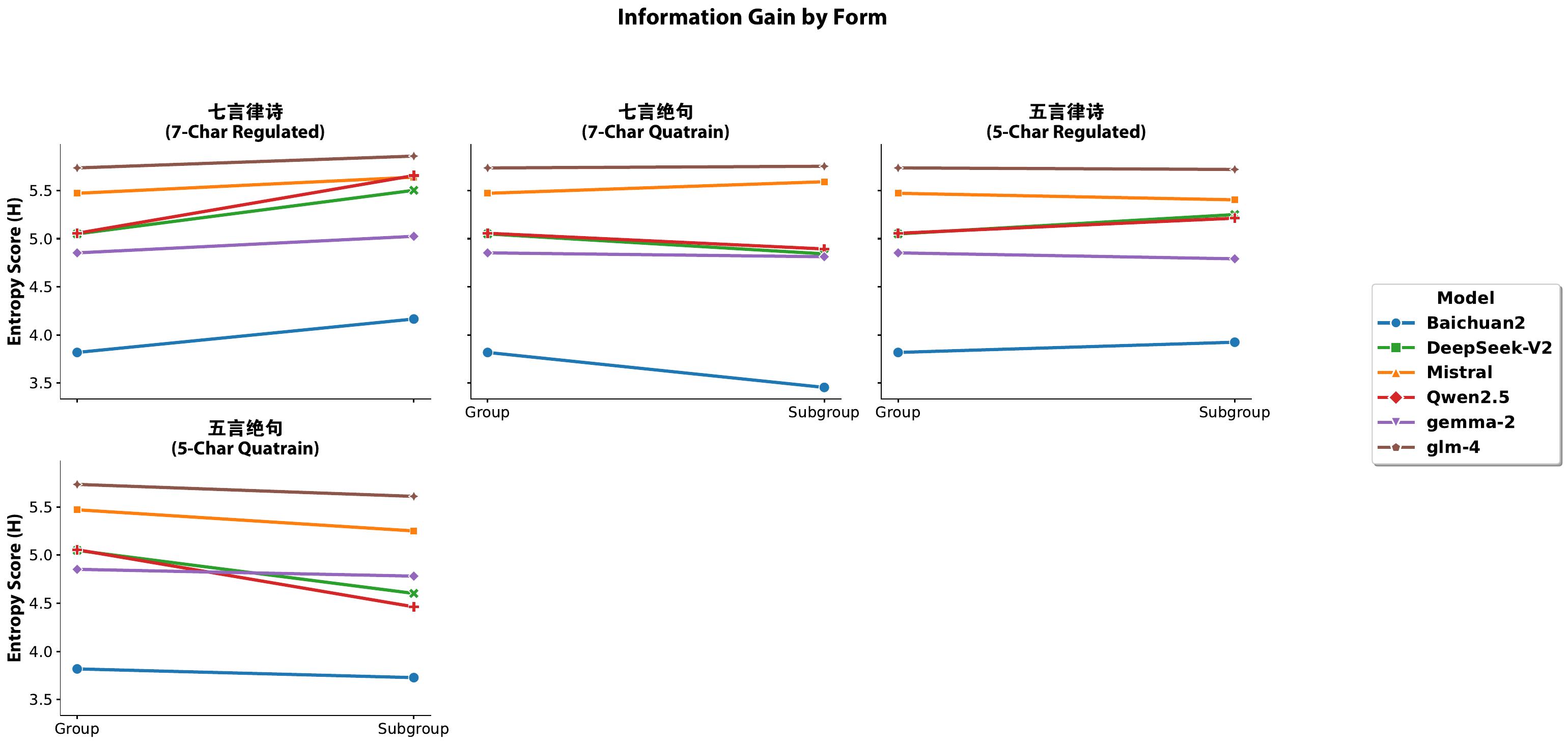}
    \caption{Information Gain by Form. Each subplot compares the overall group entropy ($H(Y)$, labeled 'Group') with the subgroup entropy conditioned on a specific poetic form ($H(Y|X_i)$, labeled 'Subgroup').}
    \label{fig:info_form}
\end{figure*}

\textbf{Poetic Form as the Strongest Constraint.} 
Across all five dimensions, Poetic Form provides the highest information gain for nearly all models (see Figure \ref{fig:info_form}). The highly structured forms, such as ``7-Char Quatrain'' and ``5-Char Regulated Verse'', consistently cause the most significant drop in entropy. This is a logical finding, as the strict rules of meter, tone, and rhyme inherent in these forms naturally limit the permissible word choices, thereby reducing the diversity of the output. This demonstrates that all models, to varying degrees, are sensitive to the explicit structural rules of classical poetry.

\begin{figure*}[htbp]
    \centering
    \includegraphics[width=1\linewidth]{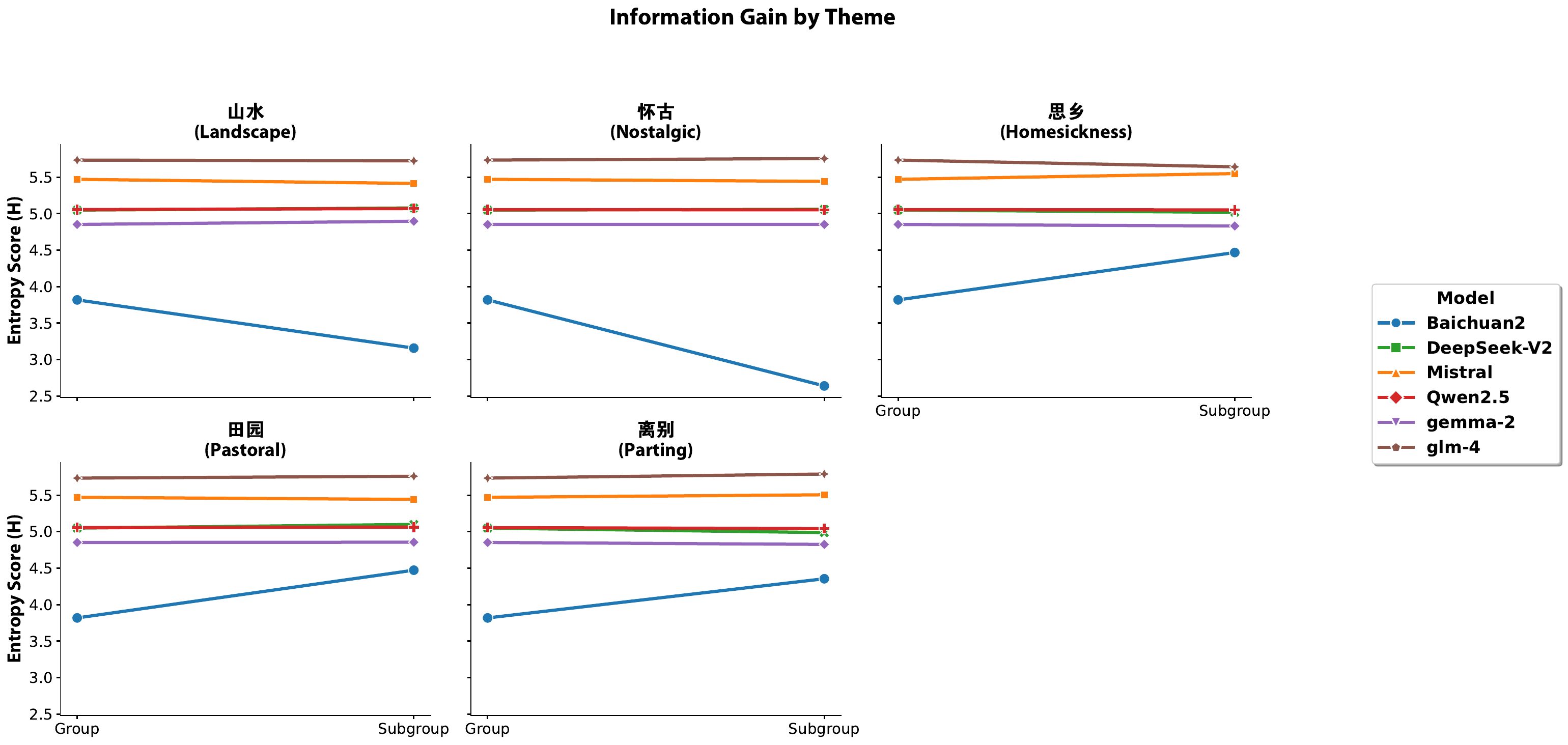}
    \caption{Information Gain by Theme. Each subplot compares the overall group entropy ($H(Y)$, labeled 'Group') with the subgroup entropy conditioned on a specific theme ($H(Y|X_i)$, labeled 'Subgroup').}
    \label{fig:info_theme}
\end{figure*}

\textbf{Thematic Prompts Provide Evidence for Baichuan's Template-Driven Strategy.} 
The analysis of the Theme dimension provides the strongest quantitative evidence for the ``template-driven'' hypothesis regarding the Baichuan model. When tasked with a specific theme like \begin{CJK}{UTF8}{gbsn}怀古\end{CJK} (``Nostalgic''), Baichuan's entropy plummets, showing by far the largest information gain of any model (see Figure \ref{fig:info_theme}). This aligns perfectly with its TF-IDF fingerprint, which showed a preference for grand, historical terms. This suggests that the ``Historical'' prompt triggers a rigid, pre-set template in the Baichuan model, leading to a highly predictable and low-diversity output.

\begin{figure*}[htbp]
    \centering
    \includegraphics[width=1\linewidth]{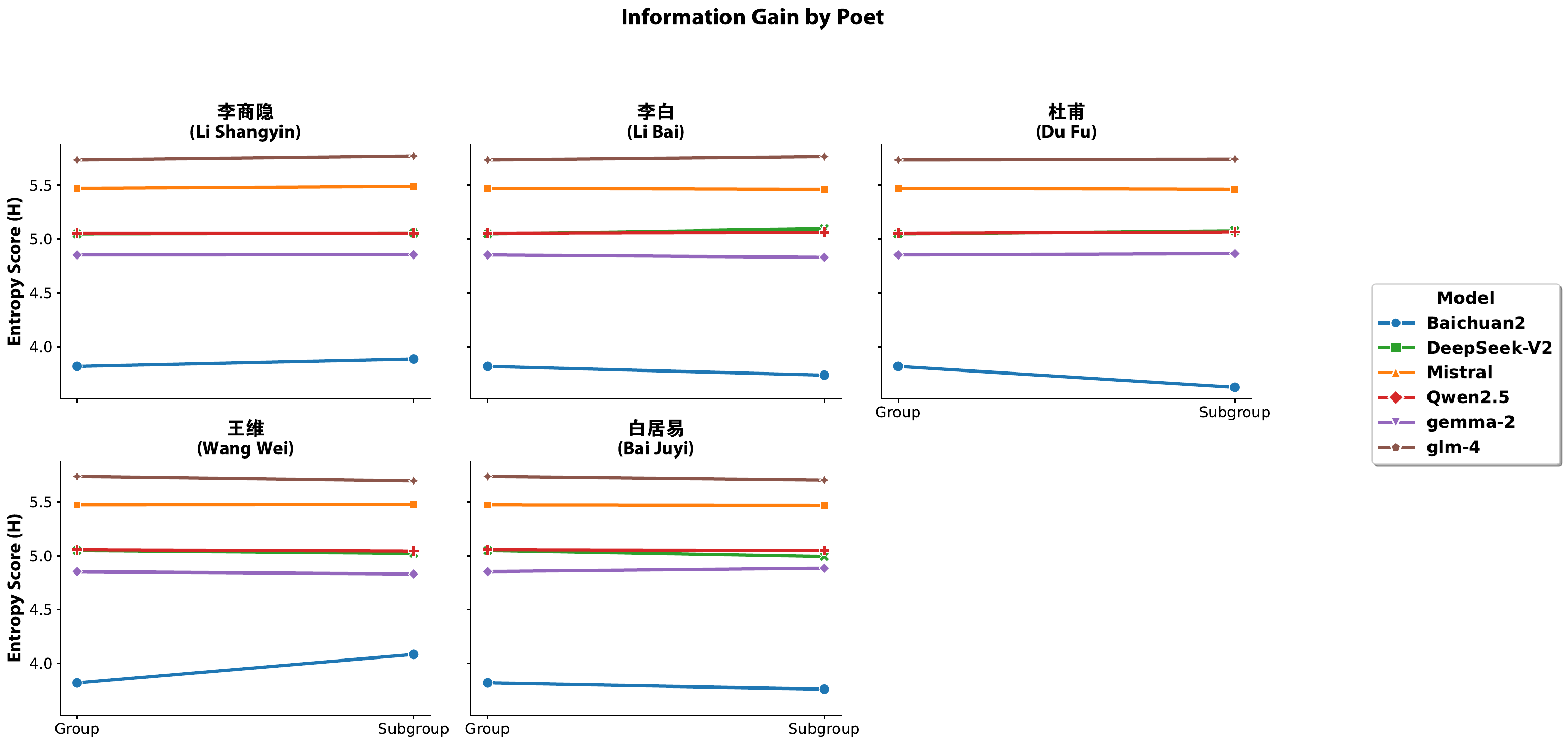}
    \caption{Information Gain by Poet. Each subplot compares the overall group entropy ($H(Y)$, labeled 'Group') with the subgroup entropy conditioned on a specific poet style ($H(Y|X_i)$, labeled 'Subgroup').}
    \label{fig:info_poet}
\end{figure*}

\begin{figure*}[htbp]
    \centering
    \includegraphics[width=1\linewidth]{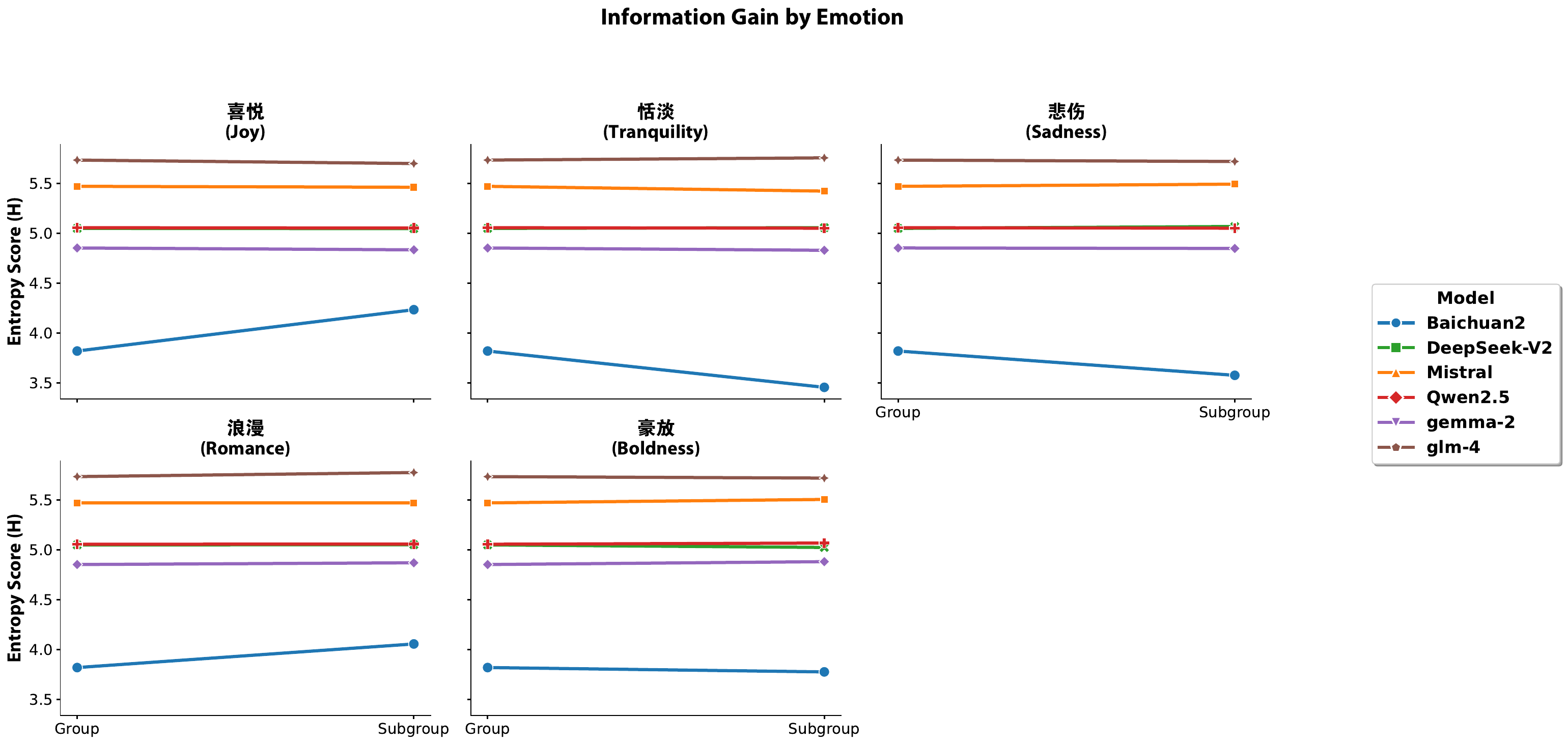}
    \caption{Information Gain by Emotion. Each subplot compares the overall group entropy ($H(Y)$, labeled 'Group') with the subgroup entropy conditioned on a specific emotion ($H(Y|X_i)$, labeled 'Subgroup').}
    \label{fig:info_emotion}
\end{figure*}

\begin{figure*}[htbp]
    \centering
    \includegraphics[width=1\linewidth]{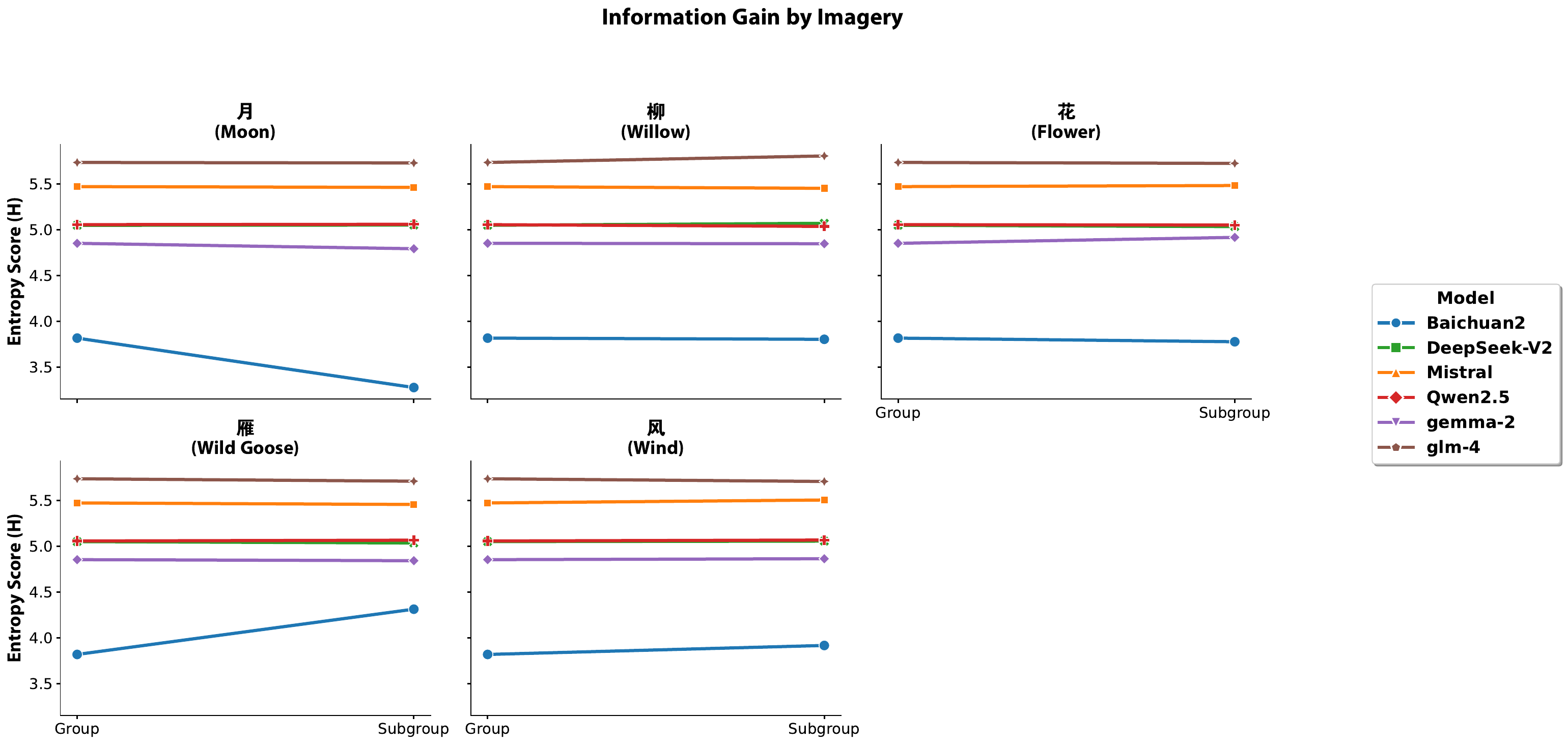}
    \caption{Information Gain by Imagery. Each subplot compares the overall group entropy ($H(Y)$, labeled 'Group') with the subgroup entropy conditioned on a specific core image ($H(Y|X_i)$, labeled 'Subgroup').}
    \label{fig:info_imagery}
\end{figure*}

\textbf{Minimal Impact from Poet, Emotion, and Imagery Dimensions.} 
In stark contrast to Poetic Form, the dimensions of Poet Style, Emotion, and Imagery show minimal information gain across most models (see Figure \ref{fig:info_poet}, Figure \ref{fig:info_emotion}, Figure \ref{fig:info_imagery}). The entropy lines for these dimensions are generally flat, indicating that specifying a poet like ``Li Bai'', an emotion like ``Sadness'', or an image like ``Moon'' does not significantly constrain the models' vocabulary. This powerful finding corroborates the results from our semantic distinction analysis (Section \ref{sec:semantic and cultural coherence}), which showed that the models struggle to produce semantically distinct styles. They may be generating poems about Li Bai or sadness, but they are not adopting a specific, constrained vocabulary that reflects a deep understanding of that style or emotion.

\textbf{The Baichuan Anomaly: An Indicator of Extreme Rigidity.} 
An interesting anomaly is observed with the Baichuan model, where for certain prompts like the theme ``Homesickness'' (\begin{CJK}{UTF8}{gbsn}思乡\end{CJK}) or the imagery ``Wild Goose'' (\begin{CJK}{UTF8}{gbsn}雁\end{CJK}), the entropy paradoxically increases for the subgroup. A plausible explanation is that Baichuan's general-purpose (Group) vocabulary is so exceptionally limited that a specific creative prompt forces it to access a different, slightly more varied set of words, thus slightly increasing its entropy relative to its own baseline. This counter-intuitive result further underscores the model's extreme rigidity and non-flexible generative process.

\section{Additional Results -- LLM Evaluation}
\label{sec:additional results llm}

\textbf{Comprehensive Evaluator Profiles: Applying the Enhanced Framework.} 
To better understand the dynamics of the LLM-as-a-judge evaluation, we analyzed the behavior of each LLM as an evaluator. This revealed four distinct evaluator profiles, each with a unique temperament and scoring bias. Mistral, for instance, acts as a ``Generous Optimist'', while Qwen behaves as a ``Strict Perfectionist''. A full comparative breakdown of these profiles, detailing their absolute and relative biases, is presented in Table \ref{tab:profile-combined}.

\section{Additional Results -- Human Evaluation}
\label{sec:additional results human}

\begin{figure*}
    \centering
    \includegraphics[width=1\linewidth]{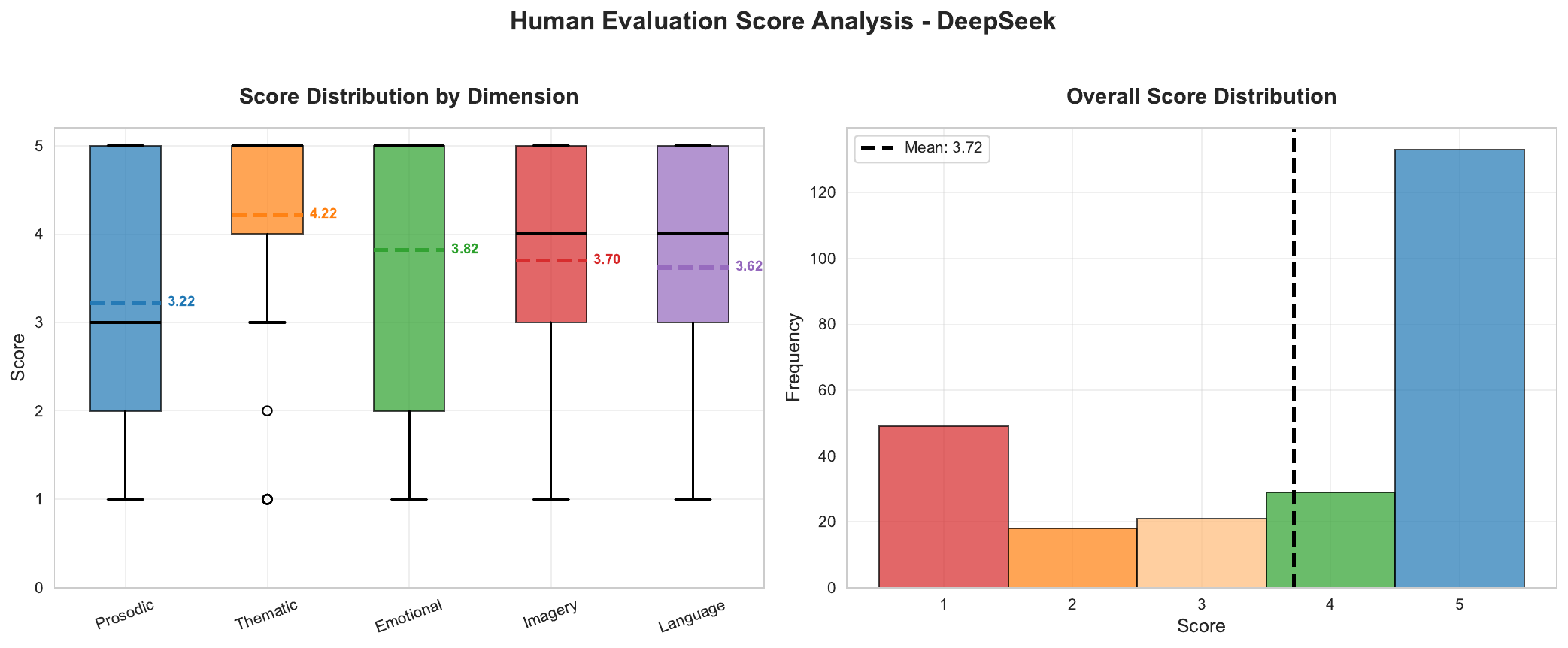}
    \caption{Human evaluation score analysis for Deepseek.}
    \label{fig:humaneval_deepseek}
\end{figure*}

\begin{figure*}
    \centering
    \includegraphics[width=1\linewidth]{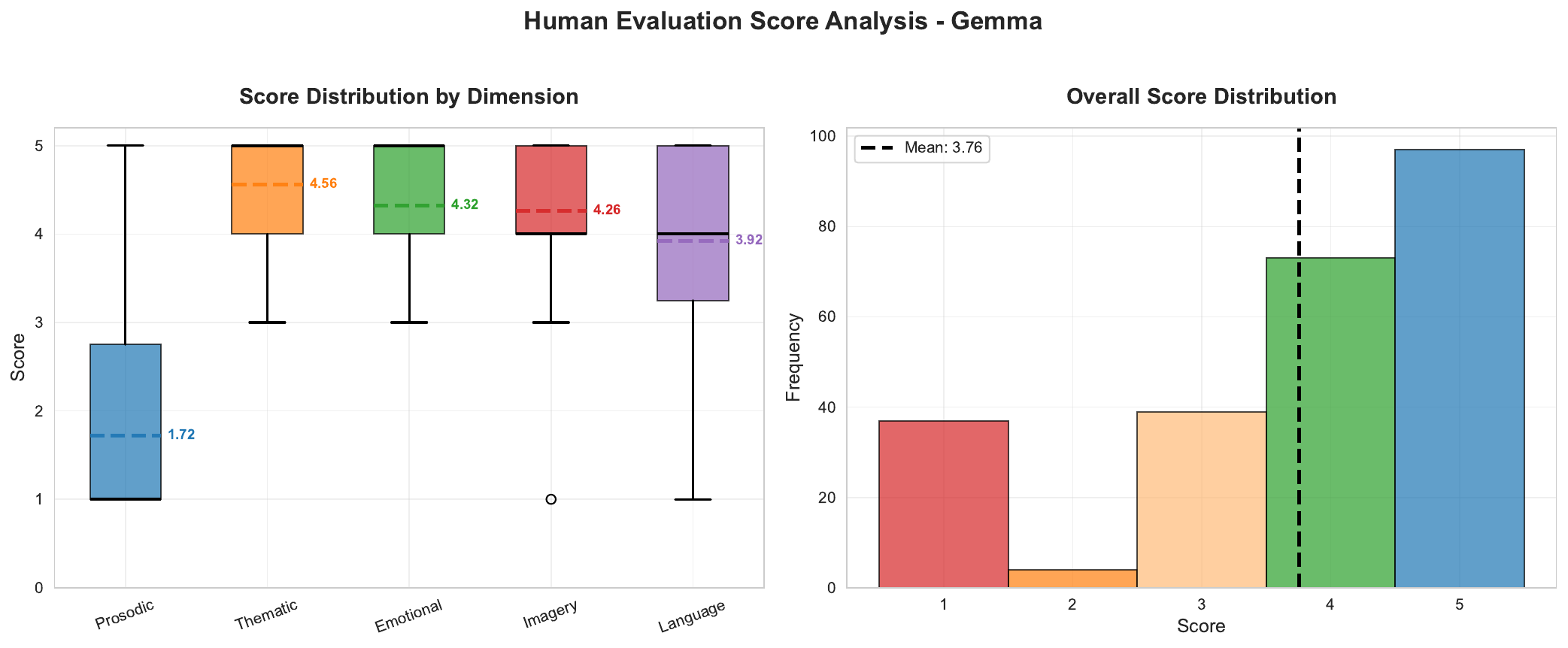}
    \caption{Human evaluation score analysis for Gemma.}
    \label{fig:humaneval_gemma}
\end{figure*}

\begin{figure*}
    \centering
    \includegraphics[width=1\linewidth]{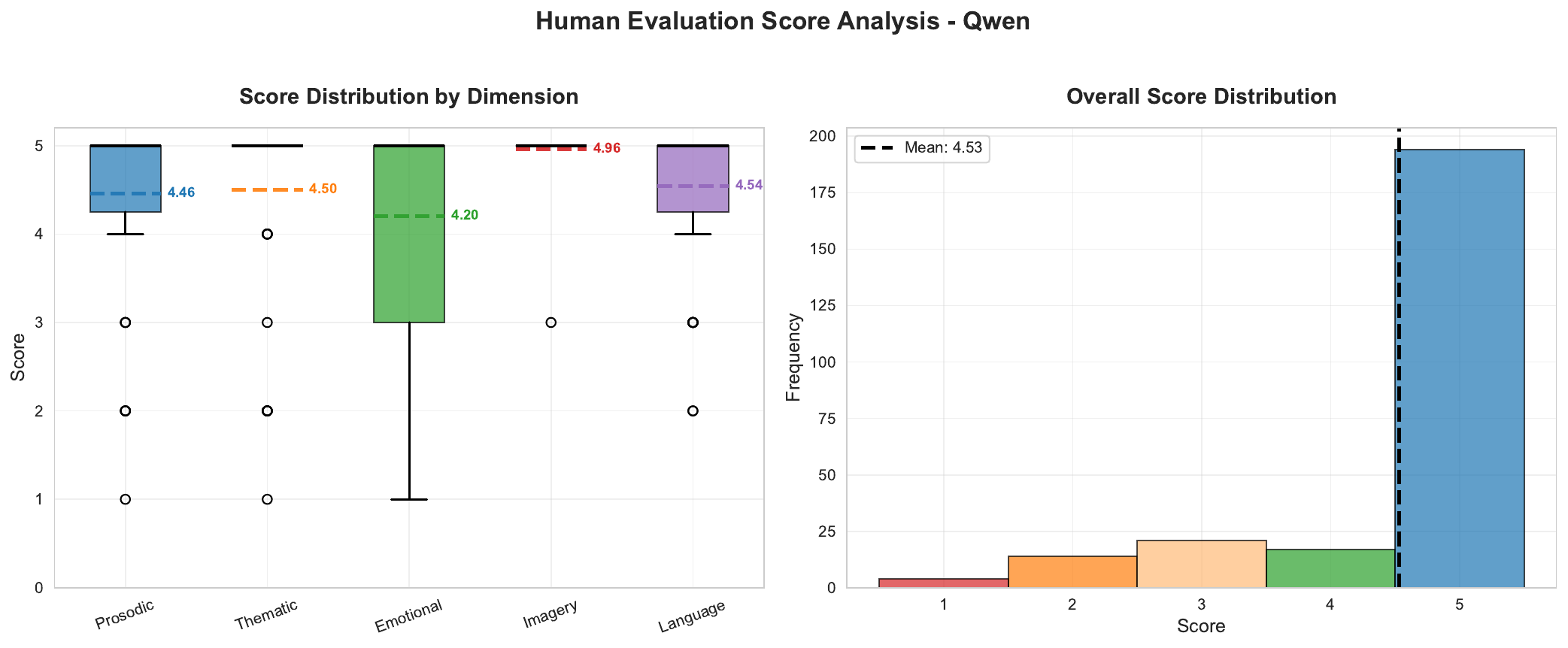}
    \caption{Human evaluation score analysis for Qwen.}
    \label{fig:humaneval_qwen}
\end{figure*}

Here we show the additional results of the overall score distribution and distributions by the five dimensions of human evaluations on the sampled subset in Figure \ref{fig:humaneval_deepseek}, \ref{fig:humaneval_gemma}, \ref{fig:humaneval_qwen}. We notice that all three models were evaluated with moderately high scores, with Deepseek and Gemma lying slightly behind Qwen. We especially notice that for both Deepseek and Gemma models, the score for prosodic adherence is largely lower than average, indicating that the models might be less capable of the prosodic nuance when generating the poems. 

\begin{figure*}
    \centering
    \includegraphics[width=0.9\linewidth]{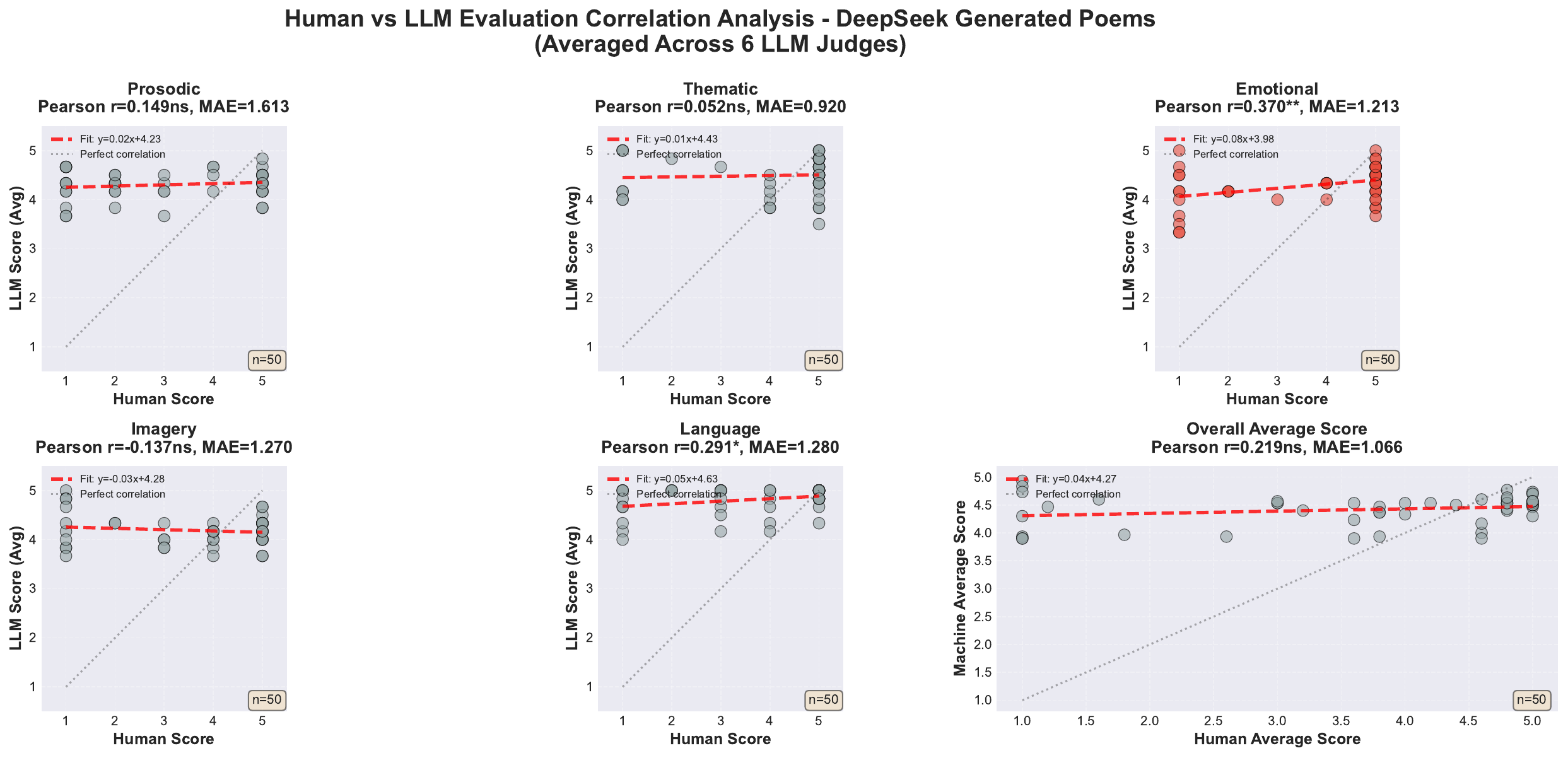}
    \caption{Human vs LLM evaluation correlation for DeepSeek generated poems (averaged across 6 LLM judges).}
    \label{fig:humanllm_deepseek}
\end{figure*}

\begin{figure*}
    \centering
    \includegraphics[width=0.9\linewidth]{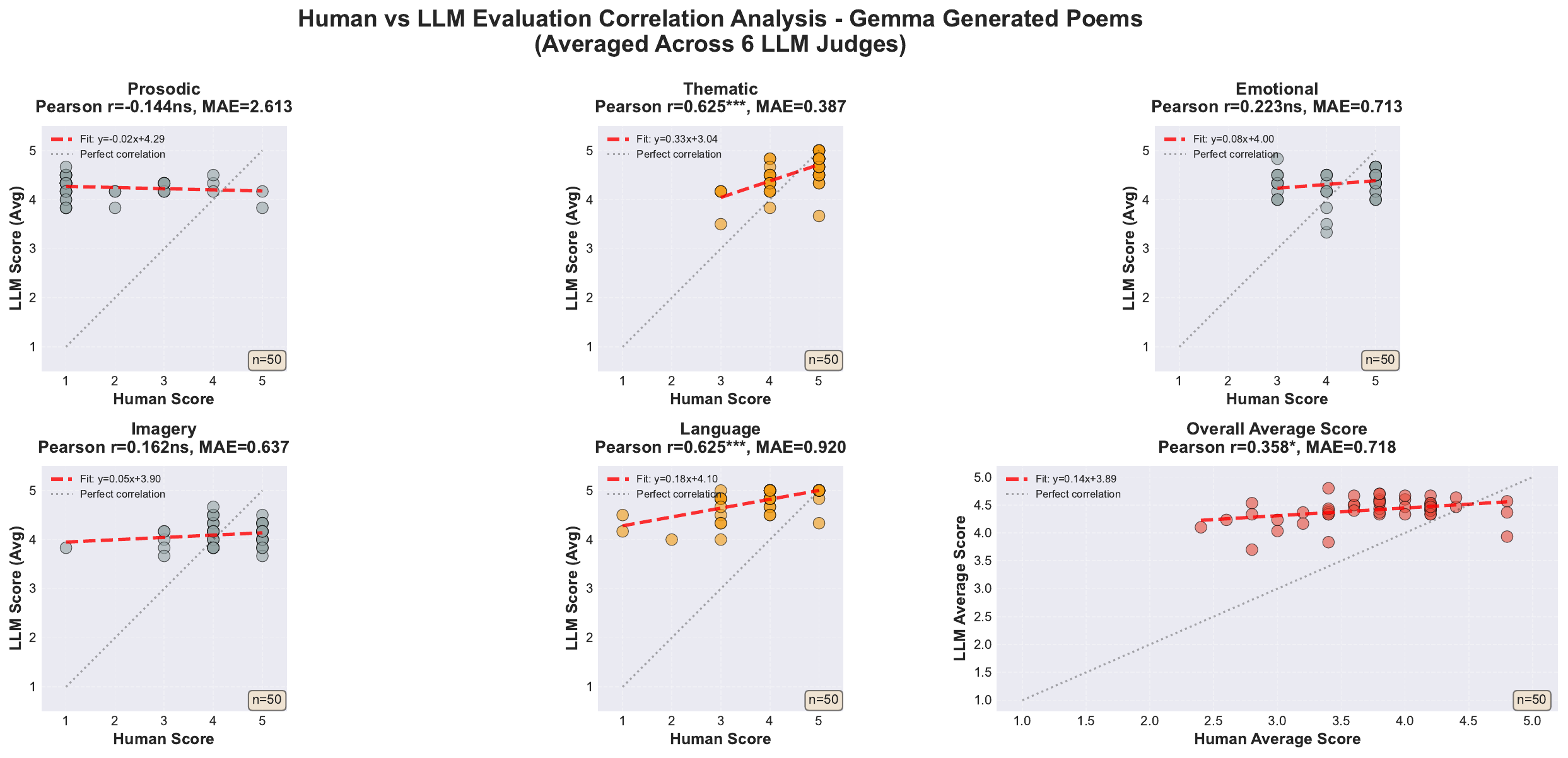}
    \caption{Human vs LLM evaluation correlation for Gemma generated poems (averaged across 6 LLM judges).}
    \label{fig:humanllm_gemma}
\end{figure*}

\begin{figure*}
    \centering
    \includegraphics[width=0.9\linewidth]{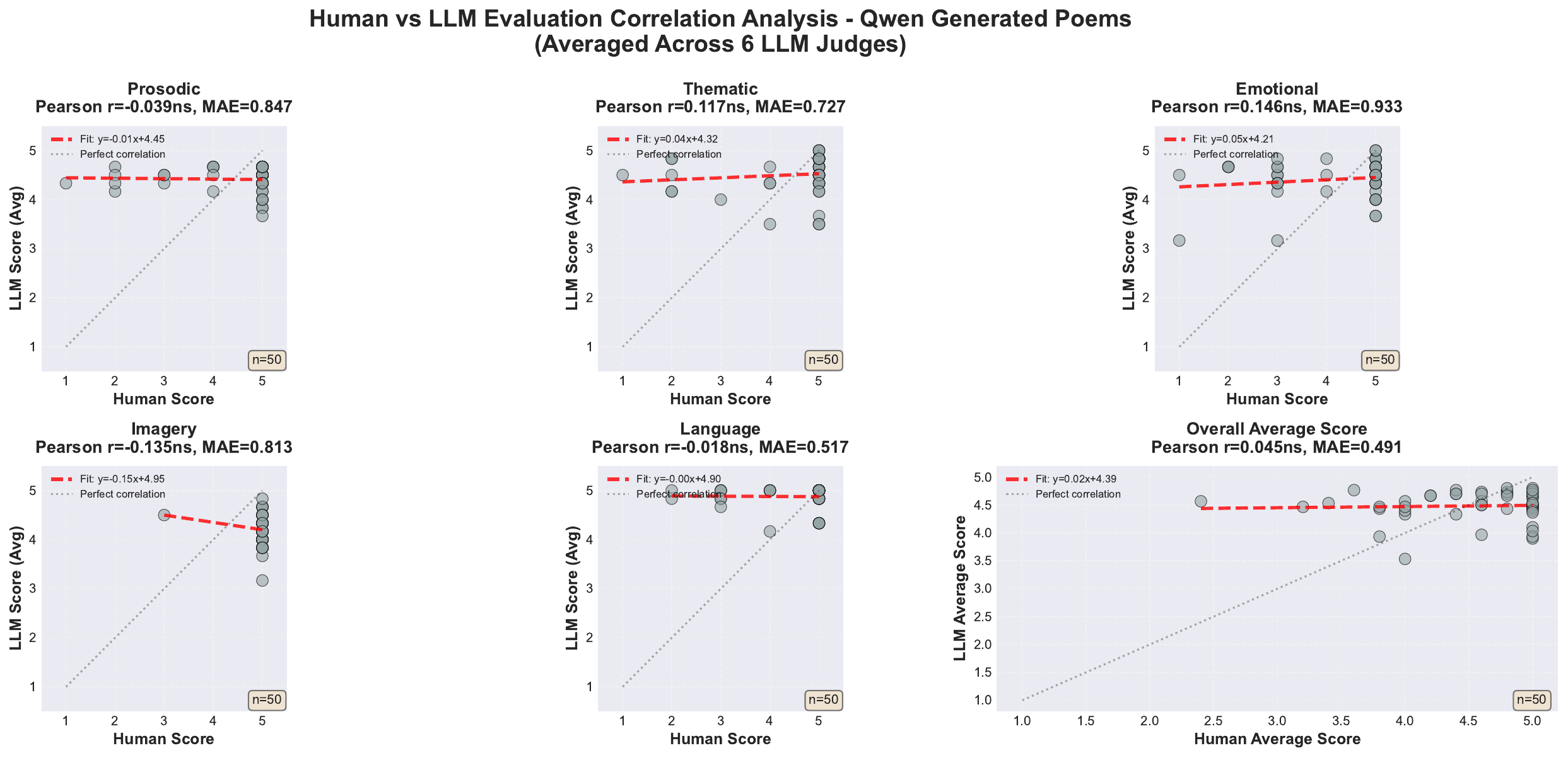}
    \caption{Human vs LLM evaluation correlation for Qwen generated poems (averaged across 6 LLM judges).}
    \label{fig:humanllm_qwen}
\end{figure*}

\begin{figure*}
    \centering
    \includegraphics[width=1\linewidth]{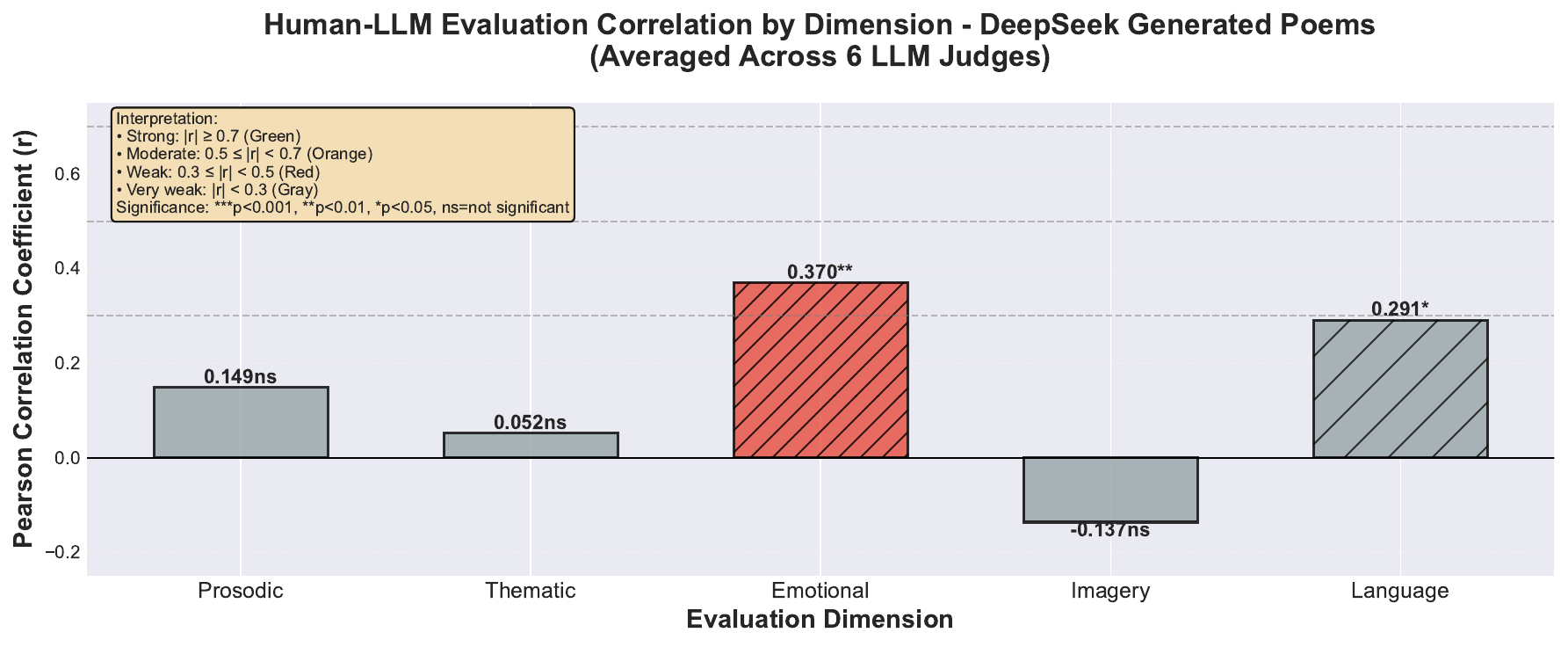}
    \caption{Human vs LLM evaluation correlation by dimension for DeepSeek generated poems (averaged across 6 LLM judges).}
    \label{fig:humandimension_deepseek}
\end{figure*}

\begin{figure*}
    \centering
    \includegraphics[width=1\linewidth]{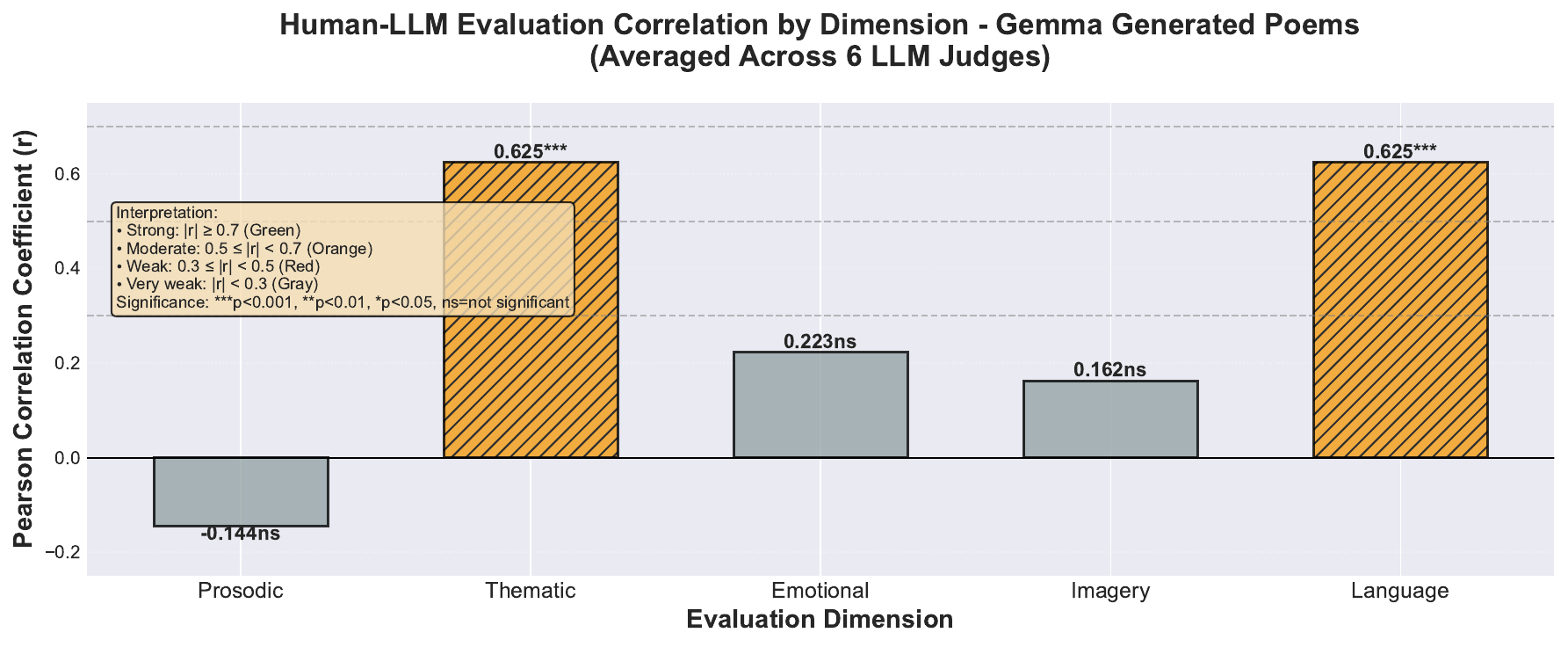}
    \caption{Human vs LLM evaluation correlation by dimension for Gemma generated poems (averaged across 6 LLM judges).}
    \label{fig:humandimension_gemma}
\end{figure*}

\begin{figure*}
    \centering
    \includegraphics[width=1\linewidth]{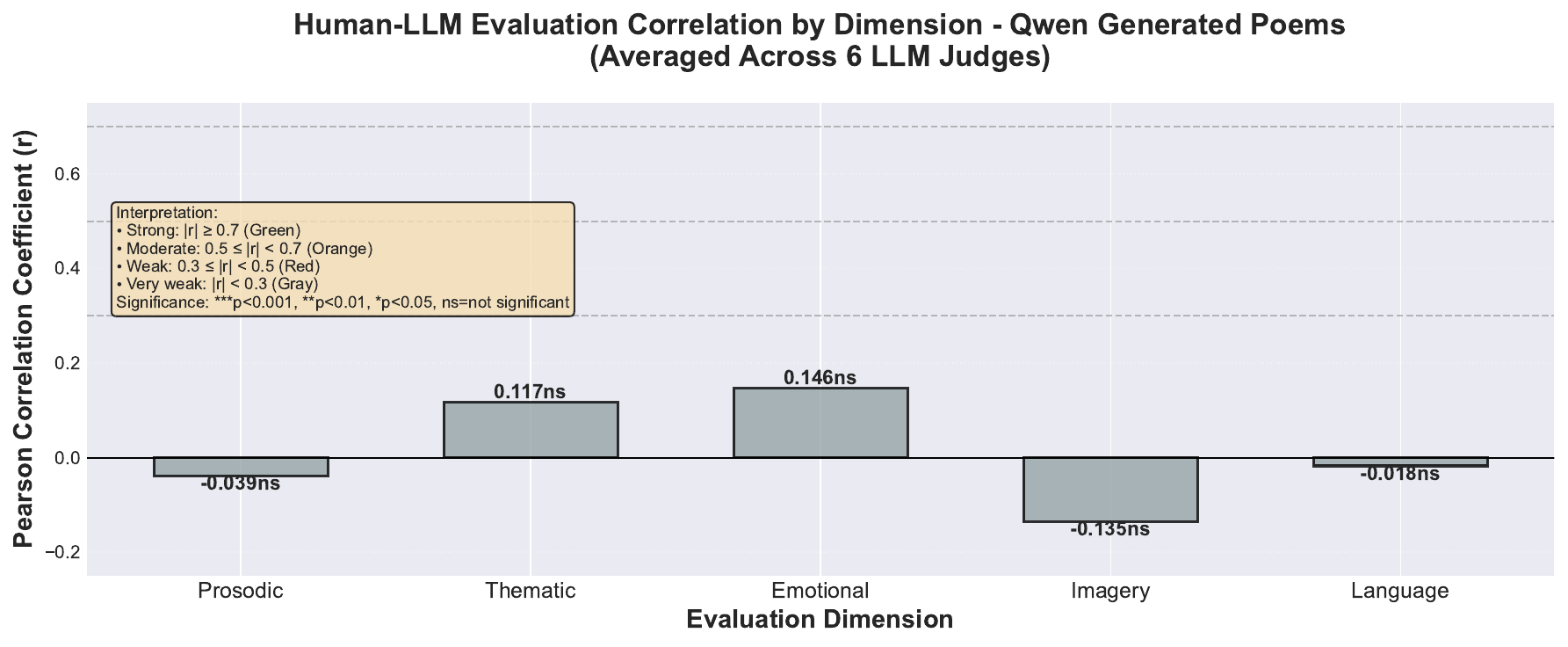}
    \caption{Human vs LLM evaluation correlation by dimension for Qwen generated poems (averaged across 6 LLM judges).}
    \label{fig:humandimension_qwen}
\end{figure*}

\begin{figure*}
    \centering
    \includegraphics[width=1\linewidth]{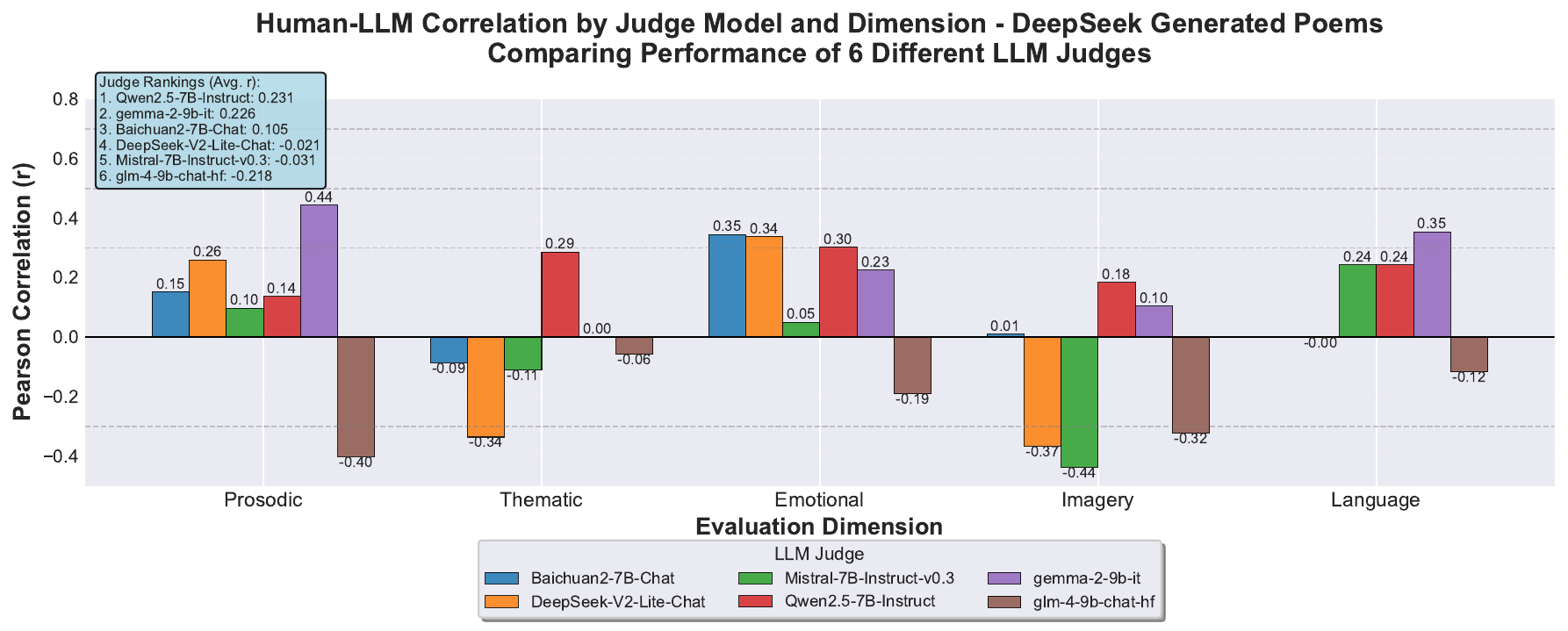}
    \caption{Human-LLM correlation by judge model and dimension for DeepSeek generated poems.}
    \label{fig:humanjudge_deepseek}
\end{figure*}

\begin{figure*}
    \centering
    \includegraphics[width=1\linewidth]{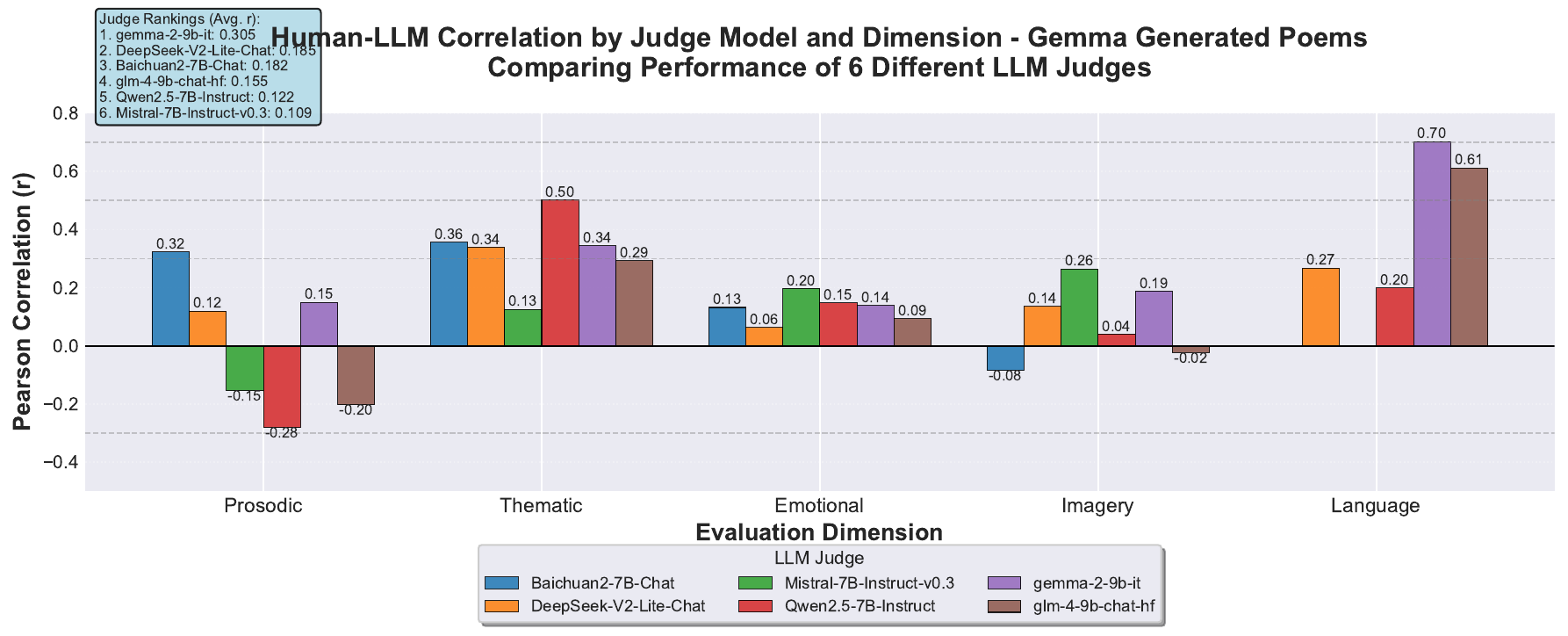}
    \caption{Human-LLM correlation by judge model and dimension for Gemma generated poems.}
    \label{fig:humanjudge_gemma}
\end{figure*}

\begin{figure*}
    \centering
    \includegraphics[width=1\linewidth]{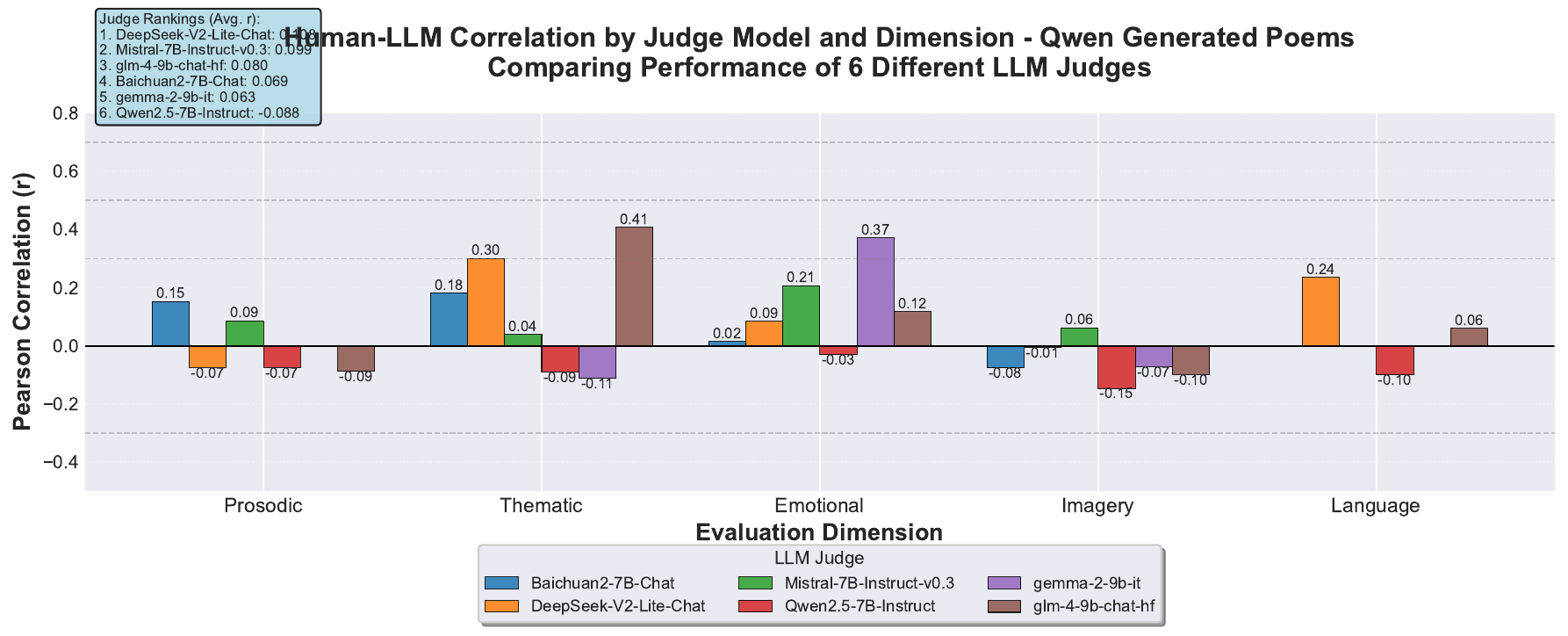}
    \caption{Human-LLM correlation by judge model and dimension for Qwen generated poems.}
    \label{fig:humanjudge_qwen}
\end{figure*}

We further did correlation analysis of the human evaluation with the LLM judges. Figure \ref{fig:humanllm_deepseek}, \ref{fig:humanllm_gemma}, \ref{fig:humanllm_qwen} visualize the correlations of human vs. LLM evaluation (averaged across 6 LLM judges) for Deepseek, Gemma and Qwen generated poems respectively. Figure \ref{fig:humandimension_deepseek}, \ref{fig:humandimension_gemma}, \ref{fig:humandimension_qwen} further visualize the the correlations of human vs. LLM evaluation by the dimension. We also show the human LLM correlation by the six judge models and different dimensions in Figure \ref{fig:humanjudge_deepseek}, \ref{fig:humanjudge_gemma}, \ref{fig:humanjudge_qwen}.


\begin{table*}[htbp]
\centering
\footnotesize
\begin{tabular}{@{}l >{\raggedright\arraybackslash}p{2.8cm} >{\raggedright\arraybackslash}p{2.8cm} >{\raggedright\arraybackslash}p{2.8cm} >{\raggedright\arraybackslash}p{2.8cm}@{}}
\toprule
\textbf{Attribute} & \textbf{Mistral} & \textbf{GLM} & \textbf{Baichuan} & \textbf{Qwen} \\
& (The Generous Optimist) & (The Balanced Self-Advocate) & (The Modest Performer) & (The Strict Perfectionist) \\
\midrule

\textbf{Temperament} & Exceptionally lenient (Avg. Score: 4.64) & Moderately lenient (Avg. Score: 4.60) & Moderately lenient (Avg. Score: 4.50) & Exceptionally strict (Avg. Score: 3.86) \\
\addlinespace 

\textbf{Absolute Bias} & +0.08 (Self: 4.72 vs. Peers: 4.64) & +0.13 (Self: 4.73 vs. Peers: 4.60) & -0.13 (Self: 4.63 vs. Peers: 4.50) & +0.20 (Self: 4.06 vs. Peers: 3.86) \\
\addlinespace

\textbf{Relative Bias} & Ranks itself 2nd (4.72) within its own scoring system. & Consistently ranks itself 1st (4.73) in its own evaluations. & Ranks itself highly but not at the top. & Ranks itself 2nd (4.06) within its own harsh framework. \\
\addlinespace

\textbf{Profile} & Confident but not arrogant. Its generosity suggests an optimistic assessment philosophy. & Demonstrates healthy self-confidence with a slight tendency toward self-promotion. & Exhibits balanced self-evaluation with a tendency toward modesty. & Presents complex behavior, maintaining high self-regard within its own stringent standards. \\

\bottomrule
\end{tabular}
\caption{\textbf{Comparative Analysis of Evaluator Model Profiles.} The table summarizes the distinct ``personalities'' of the four LLM evaluators based on their scoring behavior and self-assessment tendencies. The data highlights a clear spectrum of evaluation temperaments, from the ``Generous Optimist'' (Mistral) with an average score of 4.64, to the ``Strict Perfectionist'' (Qwen) with an average score of 3.86. Each model's absolute and relative biases in self-evaluation are also detailed.} 
\label{tab:profile-combined}
\end{table*}

\end{document}